\documentclass{article}
\pdfoutput=1

\usepackage{amssymb}

\usepackage[final]{corl_2025} 

\usepackage{etoolbox} 
\usepackage[utf8]{inputenc} 
\usepackage[T1]{fontenc}    
\usepackage{fonttable}
\usepackage[english]{babel} 
\usepackage[protrusion=true,expansion=true]{microtype}
\usepackage{comment}
\usepackage{cancel}
\usepackage{csquotes}
\usepackage{listings}
\usepackage{nameref}
\usepackage{paralist}
\usepackage{xspace}
\usepackage{setspace}
\usepackage{makeidx}
\usepackage{pdfpages}

\usepackage{times}
\usepackage{microtype}

\usepackage{amsmath,amssymb} 
\usepackage{amsfonts}       
\usepackage{mathtools}
\usepackage{array} 
\usepackage{nicefrac}       
\usepackage{breqn}          
\usepackage{textcomp}
\usepackage{bm} 
\usepackage{pifont}
\usepackage[
  separate-uncertainty = true,
  multi-part-units = repeat
]{siunitx}

\usepackage{graphicx}
\usepackage{wrapfig}
\usepackage{adjustbox} 
\usepackage{epsfig}

\usepackage{float}
\usepackage{subcaption}
\usepackage[font=small,labelfont=bf]{caption}
\usepackage{lscape}      

\usepackage{tikz}  
\usepackage{makecell}
\usepackage{overpic}
\usepackage{algorithm}
\usepackage[noend]{algpseudocode}

\usepackage{array} 
\usepackage{tabularx}
\usepackage{multirow}
\usepackage{multicol}
\usepackage{booktabs}   
\usepackage{tablefootnote}

\usepackage{paralist}
\usepackage{enumitem}
\setlist[itemize]{noitemsep,leftmargin=*,topsep=0in}
\setlist[enumerate]{noitemsep,leftmargin=*,topsep=0in}

\usepackage{url}            
\PassOptionsToPackage{table}{xcolor}
\usepackage{color,colortbl} 
\usepackage{soul}

\PassOptionsToPackage{capitalize, noabbrev}{cleveref}

\hypersetup{pagebackref=false,breaklinks=true,colorlinks,urlcolor=blue,citecolor=blue,linkcolor=blue,bookmarks=false}

\usepackage{blindtext}
\usepackage{bbm}

\usepackage{lipsum}

\usepackage{titlesec}
\titlespacing{\section}{0pt}{0.3\baselineskip}{0.25\baselineskip}
\titlespacing{\subsection}{0pt}{0.2\baselineskip}{0.15\baselineskip}
\titlespacing{\subsubsection}{0pt}{0.05\baselineskip}{0.03\baselineskip}

\renewcommand{\paragraph}[1]{\vspace{0.2em}\noindent\textit{#1} --}

\newcommand{\myalgo}{TopoCut} 

\usepackage{hyperref}

\usepackage{tikz}
\usetikzlibrary{positioning}

\title{TopoCut: Learning Multi-Step Cutting with Spectral Rewards and Discrete Diffusion Policies}

\author{
  Liquan Wang\\
  Georgia Institute of Technology \\
  \texttt{lwang831@gatech.edu} \\
  \And
  Jiangjie Bian \\
  Georgia Institute of Technology \\
  \texttt{jbian40@gatech.edu} \\
  \AND
  Eric Heidden \\
   Nvidia \\
   \texttt{eheiden@nvidia.com} \\
   \And
   Animesh Garg \\
   Georgia Institute of Technology \\
   \texttt{animesh.garg@gatech.edu} \\
}

\begin{document}
\maketitle


\begin{abstract}
Robotic manipulation tasks involving cutting deformable objects remain challenging due to complex topological behaviors, difficulties in perceiving dense object states, and the lack of efficient evaluation methods for cutting outcomes. In this paper, we introduce \textit{\myalgo}, a comprehensive benchmark for multi-step robotic cutting tasks, that integrates cutting environment and generalized policy learning. \textit{\myalgo} is built upon three core components: (1) We introduce a high-fidelity simulation environment based on a particle-based elastoplastic solver with compliant von Mises constitutive models, augmented by a novel damage-driven topology discovery mechanism that enables accurate tracking of multiple cutting pieces.
 (2) We develop a comprehensive reward design that integrates the topology discovery with a pose-invariant spectral reward model based on Laplace–Beltrami eigenanalysis, facilitating consistent and robust assessment of cutting quality. (3) We propose an integrated policy learning pipeline, where a dynamics-informed perception module predicts topological evolution and produces particle-wise, topology-aware embeddings to support PDDP—\textit{Particle-based Score-Entropy Discrete Diffusion Policy}—for goal-conditioned policy learning.
Extensive experiments demonstrate that \textit{\myalgo} supports trajectory generation, scalable learning, precise evaluation, and strong generalization across diverse object geometries, scales, poses, and cutting goals. Project page: \href{https://topocut.github.io/}{https://topocut.github.io/}.

\end{abstract}

\keywords{Deformable object manipulation, multi-step cutting, topology tracking, spectral reward, perception, discrete diffusion policy}


\section{Introduction}
Robotic manipulation involving the cutting of deformable objects plays a critical role across diverse domains such as food processing, medical surgery, and manufacturing. Many real-world tasks require not just a single cut, but a sequence of cutting actions to segment objects into complex or structured shapes. From slicing ingredients into uniform pieces in culinary automation, to performing multi-incision procedures in robotic surgery, and executing multi-pass segmentation in industrial workflows, multi-step cutting is essential for achieving fine-grained precision. The ability to reliably plan and execute these sequential cutting operations significantly enhances efficiency, safety, and quality in autonomous systems.

Despite recent progress in robotic cutting of single-material deformable objects with fixed trajectories~\cite{heiden2021disect,shi2023robocook,xu2023roboninja}, \textit{goal-conditioned multi-step cutting of complex deformable geometries} remains a major challenge. Deformable objects often fail to separate cleanly after each cut, making outcome evaluation ambiguous~\cite{heiden2021disect}. Existing evaluation metrics are sensitive to pose variations and typically require explicit alignment~\cite{xu2023roboninja}. Furthermore, dense topological changes resulting from sequential cuts are difficult to perceive from sparse or noisy observations~\cite{Xian2023,xu2023roboninja}, hindering the effectiveness of policy learning in such settings.

To address these challenges, we introduce \textit{\myalgo}, a unified framework for multi-step robotic cutting that combines high-fidelity simulation, robust evaluation, and goal-conditioned policy learning. At its core, \textit{\myalgo} features a particle-based elastoplastic simulator equipped with a novel damage-driven topology discovery mechanism that enables precise tracking of multiple cutting-induced topological changes. We further design a pose-invariant spectral reward based on Laplace–Beltrami eigenanalysis to evaluate cutting outcomes consistently across varying object geometries and poses. Finally, we propose a learning pipeline that leverages a dynamics-informed perception module to produce topology-aware, particle-wise embeddings—explicitly designed to operate on sparse visual input, making it suitable for real-world robotic settings—and supports PDDP, a discrete diffusion policy model for scalable and generalizable multi-step cutting.

Our contributions are organized into three core components:
\begin{itemize}
     \item \textbf{High-fidelity Simulation and Topology Discovery:} We develop a robust simulation environment utilizing a novel particle-based elastoplastic solver with compliant von Mises constitutive models, coupled with an advanced particle-based topology discovery method to precisely capture and track topological changes during cutting.
     
     \item \textbf{Pose-invariant Spectral Reward:} We introduce a novel reward formulation integrating the real-time topology discovery with a spectral reward function based on Laplace–Beltrami eigenanalysis, enabling consistent, pose-invariant evaluation of cutting outcomes across arbitrary object poses.

    \item \textbf{Dynamics-informed Policy Learning:} We propose a goal-conditioned policy learning framework that employs dynamics-informed perception modules to predict topology evolution and generate particle-level, topology-aware embeddings. These embeddings support conditional score-based discrete diffusion models, enhancing the robustness and generalizability of the learned cutting strategies.


\end{itemize}

\section{Related Work}

\paragraph{\textbf{Simulation Environments for Cutting}}  
Robotic cutting simulation has been explored through analytical, mesh-based, and mesh-free methods. Analytical models~\cite{Atkins2009,Zhou2006a,Zhou2006b} offer closed-form solutions limited to simple materials and motions. FEM-based methods~\cite{Areias2017EdgeCutting,Koschier2014AdaptiveFracture,Wu2015SurveyCuts} provide high-fidelity stress fields but require costly re-meshing; \textsc{DiSECt}~\cite{heiden2021disect} extends FEM with differentiable signed-distance contact and damage modeling. Mesh-free approaches like Position-Based Dynamics~\cite{Muller2007,Pan2015HybridPBD} and MPM~\cite{Stomakhin2013,Hu2018,Wang2019DuctileFractureMPM,Wolper2019CDMPM,Wolper2020AnisoMPM} better handle large topological changes. Built on MPM, \textsc{RoboNinja}~\cite{xu2023roboninja} and \textsc{FluidLab}~\cite{Xian2023} enable differentiable simulation for contact-aware and fluid–solid cutting scenarios. Our simulator extends these lines by combining Taichi-based MLS-MPM, signed-distance knife contact, and differentiable damage tracking for scalable trajectory generation.

\paragraph{\textbf{Perception for Deformable Objects}}
Deformable perception merges action and sensing to infer hidden structures~\cite{Bohg2017}. Early work fused interactive contacts for volumetric scene reconstruction~\cite{Kenney2009,Jiang2011,Schiebener2011,Schiebener2013}, while others analyzed articulation and affordances from object motion~\cite{Hausman2015,Gadre2021,Liang2022,Schmidt2019}. Multimodal sensing expanded perception to vision–touch material classification~\cite{Culbertson2014,Chu2015} and deep visuotactile pipelines~\cite{Xu2022Tandem}. Active shape reconstruction further closed the perception–action loop~\cite{Allen1990,Bierbaum2008,Matsubara2017,Xu2022Tandem3D}. In contrast, our dynamics-informed perception module predicts future topological states, producing compact task-relevant embeddings for policy learning.

\paragraph{\textbf{Deformable Object Manipulation}}  
Deformable manipulation faces challenges from high-dimensional states, occlusion, and nonlinear physics~\cite{yin2021scirob,hartmann2022long,matas2018sim2real}. Physics-centric approaches embed constitutive models~\cite{huang2020plasticinelab,lin2021diffskill,cretu2011softobj,li2021contactpoints}, while adaptive pipelines learn online corrections~\cite{chi2022iterres,matas2018sim2real,ha2022flingbot}. Imitation learning has enabled control of fluids and granular materials~\cite{seita2022toolflownet,ha2022flingbot,xie2019physical} but remains bottlenecked by data collection costs. Recently, end-to-end dynamics learning coupled with sampling-based planners or diffusion policies~\cite{shi2022robocraft,lin2022planning,driess2023compositional,chi2023diffusionpolicy} has emerged. 
We propose \textit{PDDP}, a diffusion-based policy that leverages particle-based topology-conditioned embeddings to enable goal-directed multi-step cutting of deformable objects.

\begin{figure}[t]
  \centering
  \includegraphics[width=0.95\linewidth]{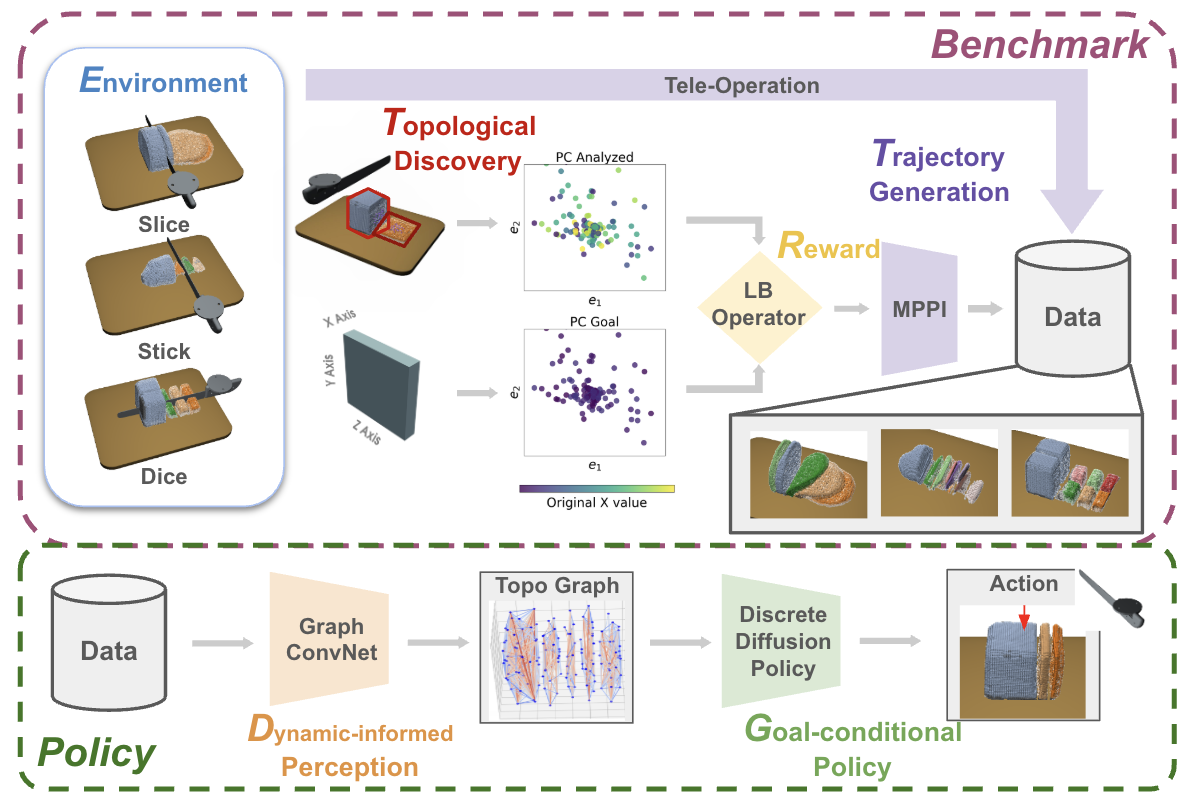}
  \caption{Overview of the \textit{\myalgo} framework: We define three representative cutting tasks—\emph{slice}, \emph{stick}, and \emph{dice}—and gather expert data via MPPI guided by Laplace–Beltrami spectral rewards and sparse tele‐operation. A dynamics‐informed perception module then extracts particle‐wise, topology‐aware embeddings from this data, which condition our PDDP to generate multi‐step cutting actions.}
  \label{fig:teaser}
\end{figure}

\section{\myalgo: Method Details}

Our goal is to enable robotic agents to segment deformable objects into goal-specified shapes through precise, sequential cutting actions. To address the challenges of topological complexity and sparse observations, we introduce \textit{\myalgo}, a unified framework that integrates simulation, perception, and control for multi-step robotic cutting. Below, we describe its key components.

\subsection{High-fidelity Simulation and Topology Discovery} 

\paragraph{Simulation Environment}
We build our deformable object cutting environment by extending FluidLab~\cite{Xian2023}, leveraging its Moving Least Squares Material Point Method (MLS-MPM)~\cite{hu2018moving} implementation in Taichi~\cite{hu2020difftaichi}. MPM is a hybrid particle-grid simulation algorithm that models continuum materials with high fidelity, making it well-suited for capturing large deformations and complex material behaviors. Our simulator augments FluidLab with several cutting-specific features, including compression/stretch-based damage tracking, von Mises plasticity with progressive softening, and surface adherence modeling. Objects are typically modeled as multi-material bodies, combining a dense plasto-elastic core with a softer von Mises outer skin to support simultaneous fracture and flow. A six-degree-of-freedom knife agent executes precise cutting actions, and we provide a teleoperation interface for intuitive control via mouse and keyboard. This extensible environment serves as the foundation for scalable data collection, evaluation, and learning in multi-step robotic cutting. For further details about MPM, please refer to Appendix~\ref{app:mpm}.

\begin{figure}[!t]
    \centering
    \begin{tikzpicture}[every node/.style={inner sep=0, outer sep=0}]
        \node[anchor=north west] (Aimg) at (0,0)
            {\includegraphics[width=0.23\linewidth]{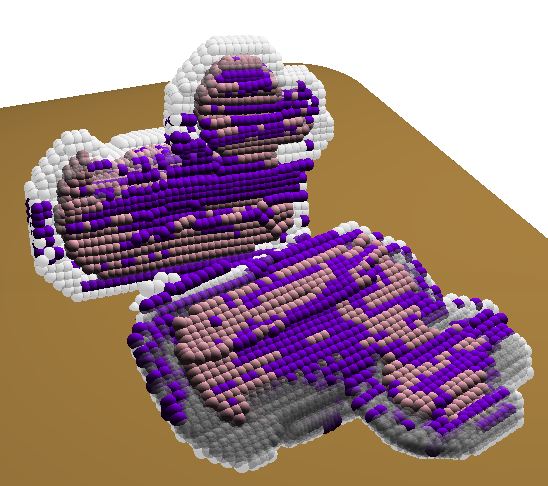}};
        \node[anchor=north west] (Acap) at ([yshift=0.15em]Aimg.south west) {%
            \begin{minipage}{0.23\linewidth}
            \vspace{0.4em}
                \small\raggedright
                \textbf{(a)} Particle-based damage tracking for topology reconstruction.
                Damaged particles (purple) mark the cut interface, from which we recover
                object segments via implicit SDF and Marching Cubes.
            \end{minipage}
        };

        \node[anchor=north west] (Cimg) at ([xshift=0.4cm]Aimg.north east)
            {\includegraphics[width=0.75\linewidth]{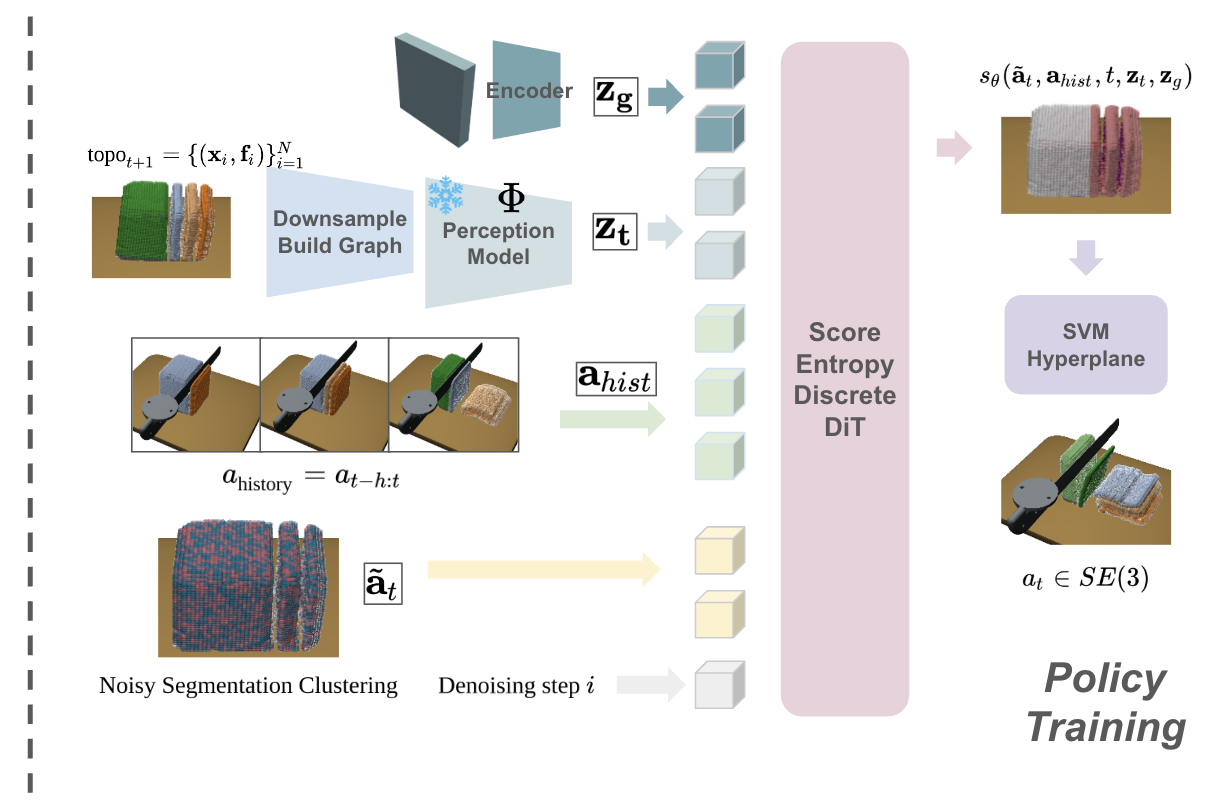}};
        \node[anchor=north west, fill=white, inner sep=1pt,
              xshift=0.58em, yshift=-0.35em] at (Cimg.north west)
            {\small \textbf{(c)}};

        \node[anchor=north west] (Bimg) at ([yshift=0.14cm]Acap.south west |- Cimg.south west)
            {\includegraphics[width=0.98\linewidth]{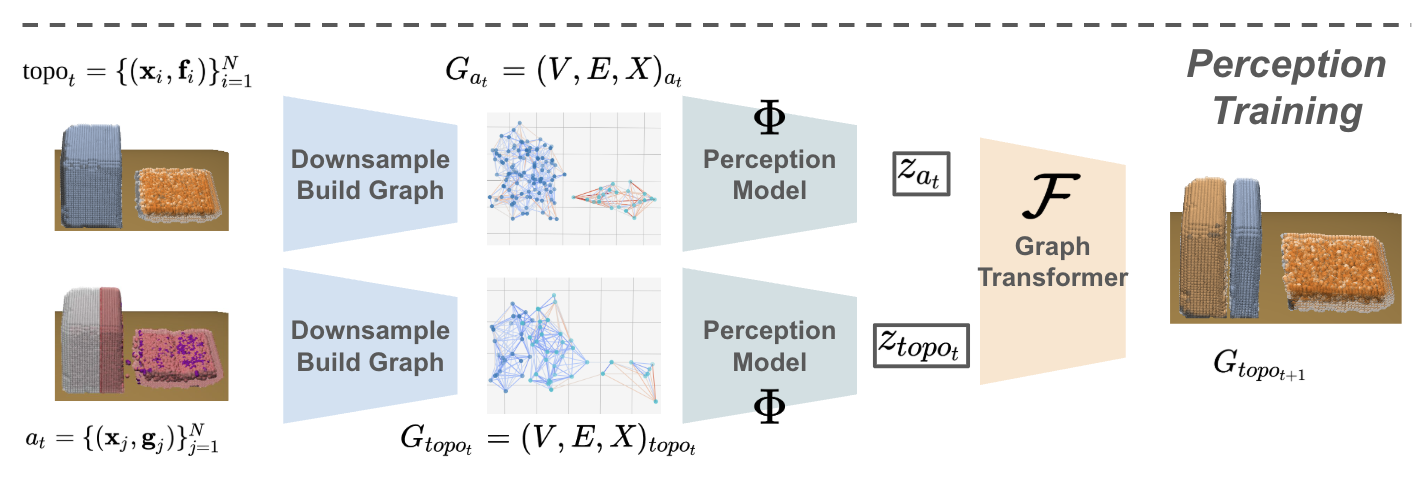}};
        \node[anchor=north west] at ([yshift=0.15em]Bimg.south west)
            {\textbf{(b)}};
    \end{tikzpicture}

    \vspace{0.1em}
    \caption{
        \textbf{(b)} Dynamics‐informed perception: given the current topological state
        $\mathrm{topo}_t$ (top left) and the cutting‐surface action $a_t$ (bottom left),
        $\mathcal{F}$ predicts the next topological state $\mathrm{topo}_{t+1}$ (right).
        A perception encoder $\Phi$ then produces particle‐wise embeddings for downstream
        policy learning.\quad
        \textbf{(c)} Particle-based score-entropy discrete diffusion policy: given the
        current topological embedding $\mathbf{z}_t$, goal embedding $\mathbf{z}_g$,
        action history $a_{\text{hist}}$, and a noised action $\tilde{\mathbf{a}}_t$,
        the policy predicts denoised cutting actions via conditional discrete diffusion.
    }
    \label{fig:combined_figures}
\end{figure}

\paragraph{Particle-Based Damage Tracking and Topology Discovery}
Accurately detecting whether a cutting action successfully separates an object is critical for evaluation. We propose a particle-based method that tracks damage and reconstructs object topology within the MLS-MPM framework. During simulation, each particle's deformation gradient \(\mathbf{F}_p\) is monitored, and a particle is classified as damaged if the volumetric Jacobian satisfies
\begin{equation}
J_p = \det(\mathbf{F}_p) \leq (1 - \epsilon_c)^m \quad \text{or} \quad J_p \geq (1 + \epsilon_s)^m,
\end{equation}
where \(\epsilon_c\) and \(\epsilon_s\) are critical compression/stretch thresholds and \(m\) is a material sensitivity parameter. For von Mises materials, damage is also triggered when yielding stress falls below a critical threshold (~\autoref{fig:combined_figures} (a)). Upon detecting damaged particles, we begin tracking the knife trajectory, which is later carved into the object's Signed Distance Field (SDF) to enforce spatial separation. The SDF is defined as
$SDF(\mathbf{x}) = \min_{p \in \mathcal{P}} \|\mathbf{x} - \mathbf{x}_p\| - r_p$, 
where \(r_p\) is the particle influence radius. 

We extract the zero-iso-surface using Marching Cubes~\cite{marchingcube}, apply Laplacian smoothing, and segment the mesh into connected components. Particles are assigned cluster identities based on proximity to reconstructed surfaces, maintaining an explicit, consistent topological representation throughout simulation. This topology discovery process provides essential signals for downstream reward computation and policy learning. For further details about particle-based damage tracking and topology discovery, please refer to Appendix~\ref{app:particle_damage}.

\subsection{Task Evaluation: Pose-invariant Spectral Reward}

\paragraph{Pose-Invariant Shape Evaluation via Spectral Analysis}
To robustly compare cut fragments to goal shapes without explicit alignment, we introduce a spectral reward based on intrinsic shape geometry. Given a point cloud \(X\), we construct a \(k\)-nearest neighbor graph with Gaussian-weighted edges \(W_{ij} = \exp(-d_{ij}^2/\sigma^2)\), compute the degree matrix \(D\), and define the combinatorial Laplacian \(L = D - W\). Solving the eigenproblem \(L\Phi = \Lambda\Phi\) yields a spectral descriptor consisting of eigenvalues \(\{\lambda_k\}\) and eigenvectors \(\{\phi_k\}\), capturing intrinsic geometry invariant to rigid motions. Given two shapes \(X\) and \(Y\), we define the spectral distance
\begin{equation}
d_{\text{spec}}(X, Y) = \alpha \|\Lambda_X - \Lambda_Y\|_2^2 + \beta \|\Phi_X^\top \Phi_X - \Phi_Y^\top \Phi_Y\|_F^2,
\end{equation}
where \(\alpha, \beta\) are weighting factors. Rather than relying on pointwise correspondences, this spectral comparison offers a compact, isometry-invariant measure of similarity. To map the spectral distance into a reward, we use an inverse scaling:
\begin{equation}
R_{\text{spec}}(X, Y) = \kappa \cdot \max(0, C - \gamma \, d_{\text{spec}}(X, Y)),
\end{equation}
where \(\gamma\) controls reward decay, \(\kappa\) is a normalization factor, and \(C\) is the pre-defined constant. For multi-fragment objects, we compute rewards over all fragment pairs and sum them to obtain the final evaluation. This formulation yields a continuous, efficient, and pose-agnostic objective for guiding goal-conditioned robotic cutting. For details about pose-invariance, please refer to Appendix~\ref{app:pose_invar}.

\paragraph{Data Generation} We gather demonstrations from two sources: (i) MPPI\,\cite{williams2017mppi}, which samples noisy knife trajectories and re-weights them via the spectral reward, and (ii) manual 6-DoF teleoperation. Each trajectory is replayed in simulation to label particle damage and reward. Every demonstration links to a goal point cloud \(g\in\{\text{slice},\text{stick},\text{dice}\}\). We store tuples
\((\mathrm{topo}_t,\,a_t,\,\mathrm{topo}_{t+1},\,g)\), yielding a concise yet diverse dataset spanning geometries, scales, poses, and tasks. Details in Appendix~\ref{app:mppi}

\subsection{Dynamics-informed Policy Learning}

\subsubsection{Dynamics-informed Perception Model Training}
To train a perception model, we formulate the evolution of object topology under cutting as $\mathcal{F} : (\mathrm{topo}_t, a_t) \mapsto \mathrm{topo}_{t+1}$, 
where \(\mathrm{topo}_t = \{(\mathbf{x}_i, \mathbf{f}_i)\}_{i=1}^N\) and \(a_t = \{(\mathbf{x}_j, \mathbf{g}_j)\}_{j=1}^N\) are particle clouds from demonstrations, with cluster labels \(\mathbf{f}_i\) and binary segmentation masks \(\mathbf{g}_j\) in one-hot form. The masks are generated by projecting the SE(3) knife action onto a cutting plane, segmenting points into labels 0 or 1.
Both clouds are downsampled via Farthest Point Sampling (FPS), then used to construct two graphs: the topology graph \(G_{\mathrm{topo}_t}\), connecting \(k\)-nearest neighbors with the same cluster label, and the action graph \(G_{a_t}\), connecting points with the same cut mask. Each graph is independently embedded through a shared perception encoder \(\Phi\):
\[
    z_{\mathrm{topo}_t} = \Phi(G_{\mathrm{topo}_t}), \quad z_{a_t} = \Phi(G_{a_t}),
\]
capturing local topological and action-induced structures. A Graph Transformer \(\mathcal{T}\) then fuses these embeddings to predict the next topological state:
$G_{\mathrm{topo}_{t+1}} = \mathcal{T}(z_{\mathrm{topo}_t}, z_{a_t})$, 
where \(G_{\mathrm{topo}_{t+1}}\) contains updated particle coordinates and cluster labels. The overall model architecture is visualized in \autoref{fig:combined_figures}(b).

Training supervision uses two objectives: a geometric loss combining Chamfer distance (CD), Earth Mover's distance (EMD), and Hausdorff distance (HD) on point positions, and a Hungarian matching loss on cluster assignments. The total loss is
$ \mathcal{L} = \lambda_{\text{pos}} \mathcal{L}_{\text{pos}} + \lambda_{\text{topo}} \mathcal{L}_{\text{topo}}$
where \(\mathcal{L}_{\text{pos}} = \text{CD} + \text{EMD} + \text{HD}\), and \(\mathcal{L}_{\text{topo}}\) minimizes the bipartite matching cost between predicted and ground-truth cluster IDs. For further details, please refer to Appendix~\ref{app:perception}.

Notably, since $\mathcal{F}$ consumes raw depth point clouds, the same network can process real sensor data without retraining, despite being trained entirely in simulation due to the low sim2real gap in depth sensor simulation~\cite{matas2018sim2real, wu2023sim2realtransferlearningpoint}. Repeated application of $\mathcal{F}$ keeps topology estimates current, bridging the sim-to-real gap and enabling accurate closed-loop manipulation on physical robots.
Real cuts hide particle-level topology. Our model infers it recursively from depth data, reproducing the simulation’s topological embeddings so the policy transfers unchanged to a real robot.

\subsubsection{PDDP: Particle-based Score-Entropy Discrete Diffusion Policy}
With the pretrained perception encoder \(\Phi\), we train a goal-conditioned behavior cloning (BC) policy to predict cutting actions. The policy takes as input the current topological observation \(\mathrm{topo}_t = (\mathbf{X}_t, \mathbf{F}_t)\), an action history \(a_{\text{hist}} = (a_{t-h}, \dotsc, a_{t-1})\), and a goal point cloud \(g\) representing a target slice, stick, or dice configuration.
We first compute point-wise embeddings:$\mathbf{z}_t = \Phi(\mathbf{X}_t, \mathbf{F}_t)$ and $ \mathbf{z}_g = \text{Encoder}(g)$
where \(\mathbf{z}_t\) captures the current spatial and topological features, and \(\mathbf{z}_g\) encodes the desired goal shape.

Cutting actions are formulated as per-point binary labels \(\mathbf{a}_t^* \in \{0,1\}^N\), classifying each particle as cut or not. Inspired by Score-Entropy discrete diffusion~\cite{lou2024sedd}, we model action prediction as a conditional discrete diffusion process (visualized in \autoref{fig:combined_figures}(c)): clean labels \(\mathbf{a}_t^*\) are progressively noised into \(\tilde{\mathbf{a}}_t\) according to $q_t(\tilde{\mathbf{a}}_t \mid \mathbf{a}_t^*) = \text{Multinomial}(\tilde{\mathbf{a}}_t \mid \mathbf{p}_t(\mathbf{a}_t^*))$
where \(\mathbf{p}_t\) defines the noise schedule.
The policy network \(s_\theta\) is trained to predict the score function:
\begin{equation}
    s_\theta(\tilde{\mathbf{a}}_t, t, a_{\text{hist}}, \mathbf{z}_t, \mathbf{z}_g) \approx \nabla_{\tilde{\mathbf{a}}_t} \log q_t(\mathbf{a}_t^* \mid \tilde{\mathbf{a}}_t),
\end{equation}
where the denoising is conditioned on the action history, current object embedding, and goal embedding.
The policy is optimized by minimizing the Denoising Score Entropy (DSE) loss:
\begin{equation}
    \mathcal{L}_{\text{BC}}(\theta) = \mathbb{E}_{(o_t, a_{\text{hist}}, g, \mathbf{a}_t^*)} \left[ \sum_{t=1}^T \left\| s_\theta(\tilde{\mathbf{a}}_t, t, a_{\text{hist}}, \mathbf{z}_t, \mathbf{z}_g) - \nabla_{\tilde{\mathbf{a}}_t} \log q_t(\mathbf{a}_t^* \mid \tilde{\mathbf{a}}_t) \right\|_2^2 \right].
\end{equation}

After the denoising process, the predicted segmentation identifies cuttable regions. A cutting plane is then fitted to the predicted cut points using Support Vector Machine algorithm, reconstructing the robot’s knife pose \(a_t \in \text{SE}(3)\). For further details, please refer to Appendix~\ref{app:pddb}.



\section{Experiments}

\subsection{Experimental Setup}

\paragraph{Environment Setup.}
Our experiments are conducted in a high-fidelity deformable object simulator based on FluidLab, leveraging the Moving Least Squares Material Point Method (MLS-MPM) implemented in Taichi. MLS-MPM accurately simulates elastoplastic deformation and topological changes with differentiable dynamics for optimization. The simulated objects consist of two distinct material layers: a dense, plasto-elastic inner core (high stiffness parameters: $\mu = 2083.33$, $\lambda = 1388.89$; high density: $\rho = 4.0$) with explicit damage tracking based on critical compression ($2.5\times10^{-2}$) and stretch ($1.0\times10^{-2}$) thresholds, and a softer, ductile von-Mises outer skin (identical stiffness, lower density: $\rho = 1.0$, with thickness $0.01$) that accommodates plastic deformation without fracture. Cuts are performed by a thin knife moving downward parallel to its blade, stopping at the cutting board, and employing high-frequency, small-amplitude oscillations to ensure clean separation. After each cut, an optional gentle push helps fragments settle onto the board.

\paragraph{Task Setup.}
We evaluate goal-conditioned cutting tasks to segment the deformable object into predefined shapes: slices (thin planar sections with controllable thickness), sticks (obtained by further cutting slices along their width), and dices (small blocks produced by rotating the knife $90^\circ$ and chopping sticks orthogonally). Each task varies in slice thickness, stick width, and dice size. For stick and dice tasks, a gentle downward push ensures proper fragment separation between cuts. Success of each cut is determined by comparing the fragment shape to the target using our spectral reward (based on the Laplace-Beltrami operator). A predefined reward threshold is used both to evaluate successful shape matching and to terminate MPPI-based trajectory planning.

\subsection{Benchmark Evaluation}

\begin{figure}[t]
  \centering
  \begin{minipage}{0.68 \textwidth}
  \centering
  \includegraphics[width=\linewidth]{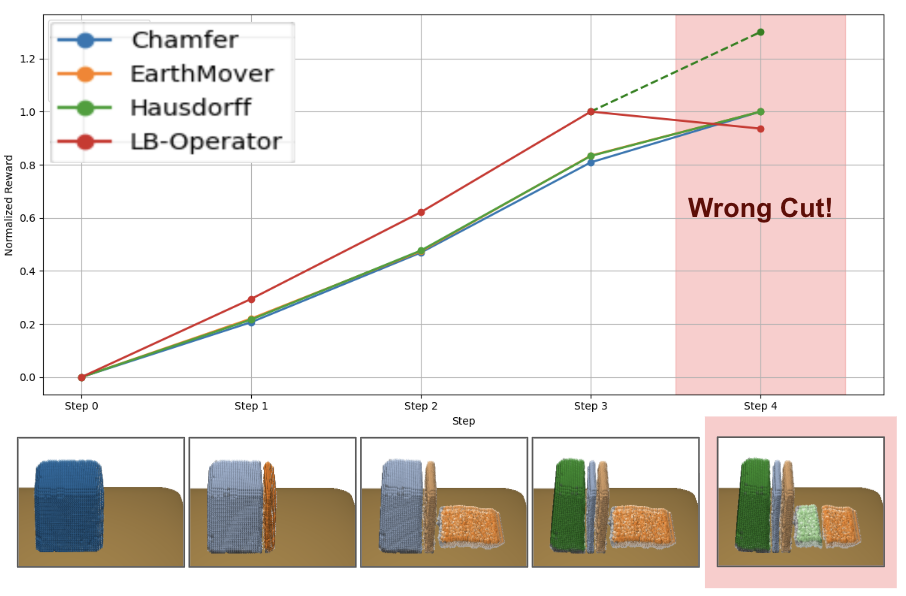} 
  \end{minipage}
  \begin{minipage}{0.31\textwidth}
  \centering
  \caption{
  \textbf{Qualitative reward evaluation via multi-step cutting}
We assess a four-step cutting sequence, comparing each piece to a target shape. The first three cuts match the goal; the last is intentionally wrong. Across metrics, only our spectral reward (red) rises during the correct cuts and drops after the failure, reflecting shape fidelity while remaining pose-invariant. Other metrics keep increasing despite the bad final cut. When the last cut is corrected, the spectral reward rises again (green dashed), confirming its robustness.
    }
  \label{fig:cut_qual}
  \end{minipage}
  \vspace{-10pt}
\end{figure}

\paragraph{Q1: Can our spectral reward reliably indicate the success or failure of sequential cutting actions?}  
To evaluate the reliability of our spectral reward, we conducted a teleoperated slicing experiment involving a four-step cutting sequence. The first three cuts produced segments that matched the target shape, while the final cut was deliberately incorrect. At each step, we computed normalized similarity scores using Chamfer Distance (CD), Earth Mover’s Distance (EMD), Hausdorff Distance (HD), and our spectral reward.
As shown in~\autoref{fig:cut_qual}, only our spectral reward steadily increases during the successful cuts and drops sharply after the failed one, correctly capturing true shape fidelity while remaining pose-invariant. In contrast, CD, EMD, and HD continue to rise even after the erroneous cut, failing to reflect the degradation in quality. When we replace the failed final cut with a correct one, the spectral reward resumes increasing (green dashed line), further confirming its robustness and sensitivity to meaningful geometric improvement.

\paragraph{Q2: Can our spectral reward support large-scale data generation for policy learning?}  
We combine our spectral reward with MPPI planning to autonomously generate expert demonstrations across three goal types—\texttt{slice} ( {\includegraphics[height=0.8\baselineskip]{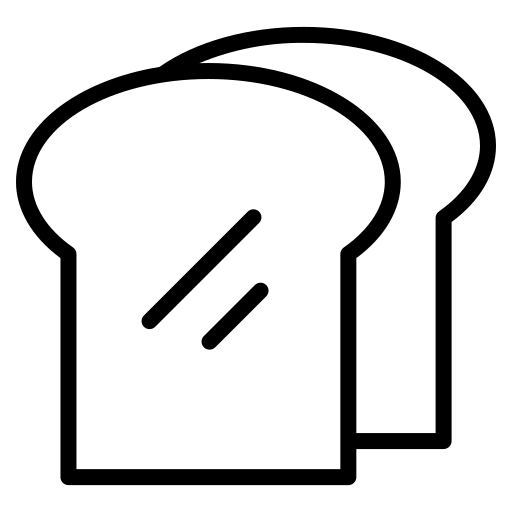}} ), \texttt{stick} ( {\includegraphics[height=0.8\baselineskip]{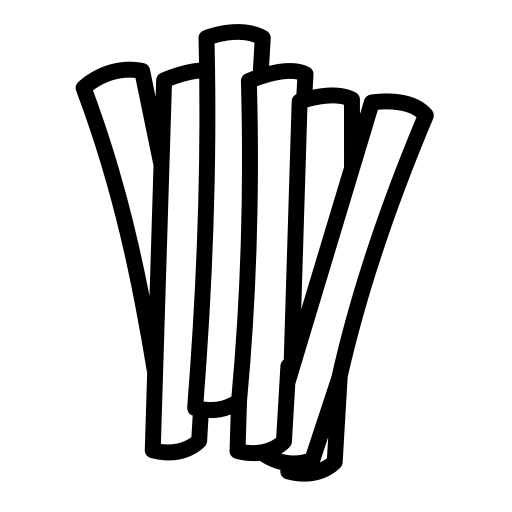}} ), and \texttt{dice} ( {\includegraphics[height=0.8\baselineskip]{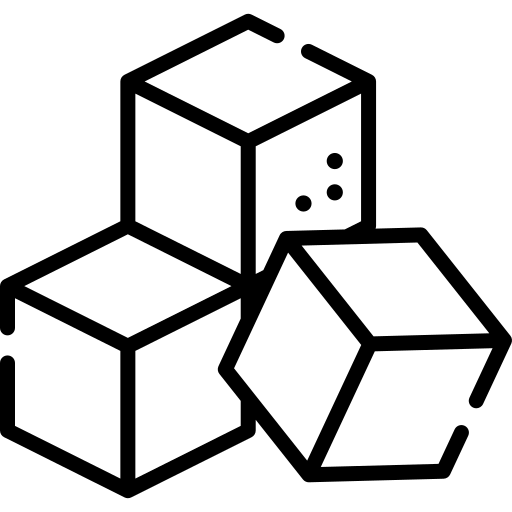}} )—each with varied specifications. Cutting is performed on 10 diverse object geometries (e.g., cake, potato, tomato), resulting in ~1,000 trajectories with around five cuts each.
This dataset captures a wide range of shapes, poses, and cut targets, and is used to train goal-conditioned cutting policies in the next stage.

\subsection{Policy Evaluation}
\begin{table*}[t] \LARGE
  \centering
  \caption{\small
  \textbf{Normalized Spectral Reward ($\hat{R}$) and Cut Count ($N_C$):}
  All $R$ values are normalized by the corresponding human baseline per cut mode. Overall, our method yields superior normalized rewards and efficient cut counts across both in-distribution and out-of-distribution geometries, demonstrating robust performance and generalization.
  }
  \label{table:cutting_eval}
  \renewcommand{\arraystretch}{1.5}
  \setlength{\tabcolsep}{10pt}
  \resizebox{0.95\linewidth}{!} {
  \begin{tabular}{l||cc|cc|cc||cc|cc|cc}
  \toprule
\rowcolor[HTML]{FFDEB4}
\textbf{} 
& \multicolumn{6}{c||}{\textbf{In-distribution Geometries}} 
& \multicolumn{6}{c}{\textbf{Out-of-distribution Geometries}} \\

\rowcolor[HTML]{FFDEB4}
\textbf{}
& \multicolumn{2}{c|}{\includegraphics[width=0.06\linewidth]{figures/slice.png}}
& \multicolumn{2}{c|}{\includegraphics[width=0.06\linewidth]{figures/stick.png}}
& \multicolumn{2}{c||}{\includegraphics[width=0.06\linewidth]{figures/cube.png}}
& \multicolumn{2}{c|}{\includegraphics[width=0.06\linewidth]{figures/slice.png}}
& \multicolumn{2}{c|}{\includegraphics[width=0.06\linewidth]{figures/stick.png}}
& \multicolumn{2}{c}{\includegraphics[width=0.06\linewidth]{figures/cube.png}} \\

\midrule

\rowcolor[HTML]{ACCEAA}
\textbf{Algorithm}
& $\hat{R}$ & $N_C$
& $\hat{R}$ & $N_C$
& $\hat{R}$ & $N_C$
& $\hat{R}$ & $N_C$
& $\hat{R}$ & $N_C$
& $\hat{R}$ & $N_C$ \\

\midrule

\rowcolor[HTML]{FFFFFF}
Human Tele-Op & 3.3 & 4.7 & 1.7 & 3.5 & 3.9 & 5.0 & - & - & - & - & - & - \\\cmidrule{1-13}
\midrule

\rowcolor[HTML]{EFEFEF}
Heuristic 
& 0.82 & 2.40 
& 0.53 & 0.90 
& \textbf{0.59} & 1.90 
& \textbf{0.88} & 2.80 
& 0.29 & 0.40 
& 0.41 & 1.60 \\\cmidrule{1-13}

\rowcolor[HTML]{FFFFFF}
DP3 
& 0.74 & 2.72 
& 0.50 & 0.92 
& 0.49 & 2.29 
& 0.85 & 2.90 
& 0.26 & 0.52 
& 0.42 & 1.60 \\\cmidrule{1-13}

\rowcolor[HTML]{EFEFEF}
3D Diffuser Actor
& 0.83 & 2.91 
& 0.66 & 1.00 
& 0.52 & 2.54 
& 0.58 & 2.42 
& 0.31 & \textbf{0.95}
& 0.43 & 1.40 \\\cmidrule{1-13}

\rowcolor[HTML]{FFFFFF}
Ours w/o Perception 
& \textbf{0.92} & \textbf{5.25} 
& 0.56 & 0.93 
& 0.51 & \textbf{2.58} 
& 0.64 & 3.22 
& 0.25 & 0.94 
& 0.45 & 1.51 \\\cmidrule{1-13}

\rowcolor[HTML]{EFEFEF}
\textbf{PDDP (Ours)} 
& 0.85 & 4.83 
& \textbf{0.71} & \textbf{1.15} 
& \textbf{0.57} & 2.33 
& 0.82 & \textbf{4.31} 
& \textbf{0.39} & 0.84 
& \textbf{0.49} & \textbf{1.89} \\

\bottomrule
\end{tabular}
}
\end{table*}

\paragraph{Evaluation Setup.}  
We evaluate policies in the MLS-MPM simulator using deterministic rollouts of five sequential cuts per episode. Each policy is tested on 405 episodes spanning nine object geometries—five seen during training (\textit{cake, cube, ditto, ham, potato}) and four novel (\textit{donut, burger, sushi, steak})—across three goal types: \texttt{slice} ( {\includegraphics[height=0.8\baselineskip]{figures/slice.png}} ), \texttt{stick} ( {\includegraphics[height=0.8\baselineskip]{figures/stick.png}} ), and \texttt{dice} ( {\includegraphics[height=0.8\baselineskip]{figures/cube.png}} ). Object pose is randomized in translation ($x \in [-0.4, 0.4]$, $z \in [-0.2, 0.2]$), yaw ($\in [-15^\circ, 15^\circ]$), and scale ($\in [0.8, 1.2]$) to assess robustness.

\paragraph{Baselines.}  
We compare our method to four baselines: a 3D diffusion model (DP3), a standard point cloud diffusion model (3D Diffuser Actor), a heuristic controller with fixed cut trajectories, and human teleoperation. Each learned policy is tested with three perception configurations: (i) no perception (raw point clouds only), (ii) a frozen pretrained graph encoder $\Phi$, and (iii) a self-attention-based GraphConv encoder. Our full model uses the policy paired with $\Phi$.

\paragraph{Evaluation Metrics.}  
We report two metrics: (1) \textit{normalized spectral reward} \( \hat{R} \), which captures cut quality via spectral similarity to the goal shape (normalized to human teleop scores); and (2) \textit{cut count} \( N_C \), the number of correctly segmented goal-consistent pieces. These metrics jointly reflect geometric accuracy and task success. Results are summarized in Table~\ref{table:cutting_eval}.



\paragraph{Q3: How does the choice of policy model and perception architecture impact cutting performance?}  
Our PDDP backbone consistently outperforms baselines across both seen and unseen tasks. Instead of directly predicting actions, it segments point-wise regions and infers actions via an SVM, improving stability and noise tolerance—achieving up to 85\% test accuracy post-training. Using a frozen, pretrained perception encoder further stabilizes learning and enhances generalization; its removal often leads to overfitting and degraded performance on unseen geometries. Many failure cases involve disorganized particle states after poor cuts, where our segmentation-based transformer helps maintain spatial consistency. For fair comparison, all diffusion baselines are configured to predict one sparse action per step, matching our PDDP setup.

\paragraph{Q4: How well can policies trained with our framework generalize across varied object geometries and goal shapes?}  
Our method generalizes well to unseen objects and goals, maintaining high spectral rewards and accurate cut counts. This is enabled by the pretrained perception module and diverse demonstrations. Without the pretrained encoder, policies overfit and perform poorly on novel shapes. The \textit{stick} task shows the lowest scores across all methods, as it depends heavily on global geometry—irregular shapes make clean stick cuts difficult. We also observe that cube-like shapes yield more valid \textit{stick} and \textit{dice} cuts than irregular or donut-like ones, especially under our strict reward threshold. These results underscore the importance of structured perception and geometry-aware learning in complex cutting.

\begin{figure}[t]
  \centering
  \includegraphics[width=\linewidth]{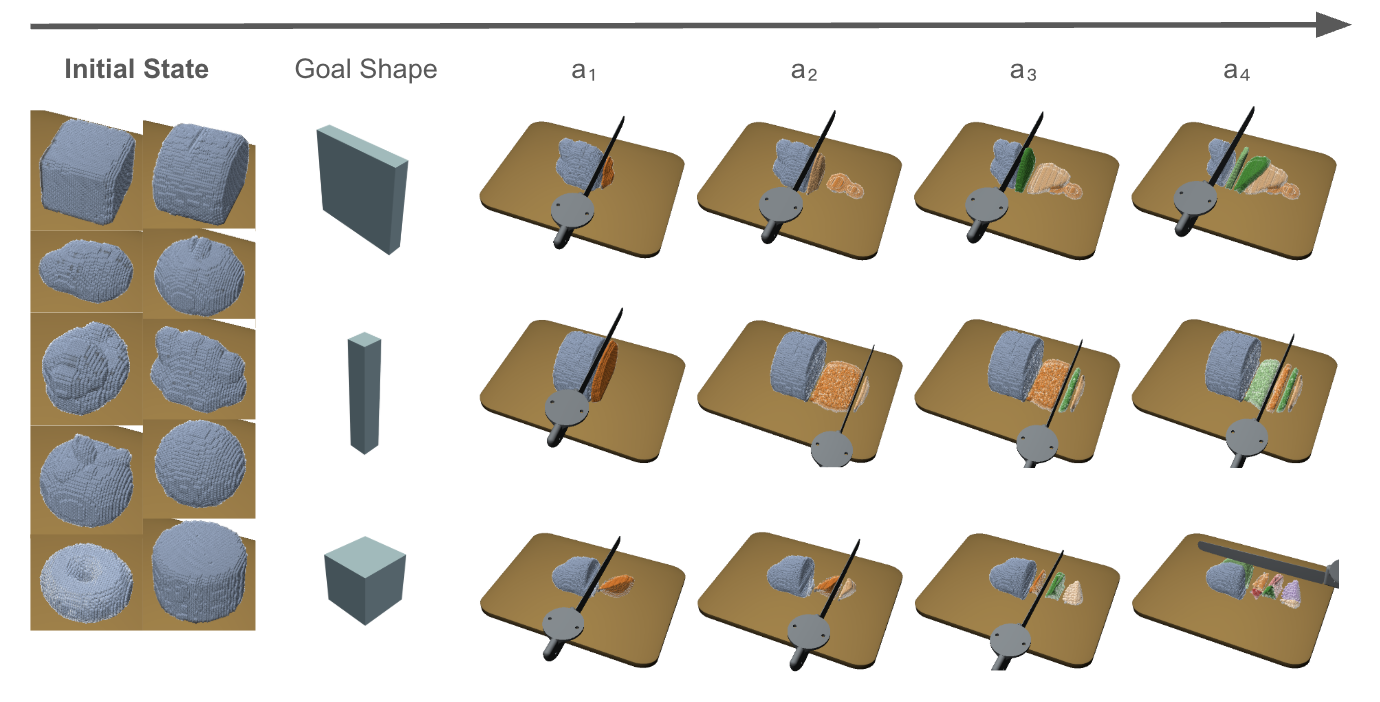}
  \caption{Example cutting trajectories for the three canonical tasks. Each row corresponds to one task—\emph{slice} (top), \emph{stick} (middle), and \emph{dice} (bottom)—showing the initial state (column 1), the goal shape (column 2), and the successive knife poses and resulting fragments at Steps 1–4 (columns 3–6).}
  \label{fig:cut_steps}
\end{figure}

\section{Conclusion}
We introduced \textit{\myalgo}, a unified platform for robotic cutting of deformable objects that combines a high-fidelity particle simulator, robust topology tracking, and a pose-invariant spectral reward for cut quality. \textit{\myalgo} trains goal-conditioned cutting policies via a conditional score-based diffusion model and a pretrained, dynamics-informed perception module capturing topological changes. Experiments show strong generalization across geometries, materials, and tasks, consistently outperforming baselines. \textit{\myalgo} sets a new benchmark for deformable object manipulation, with broad relevance to industrial, medical, and personal robotics.

\newpage
\section{Limitation}
Despite its significant advancements, \textit{\myalgo} still faces notable limitations that should be addressed in future work. In particular, the warp mesh reconstruction stage dominates the computational workload, forcing the simulator to run at real-time speeds and restricting efficient large-scale online reinforcement learning. Mitigating this bottleneck through algorithmic optimizations, GPU-accelerated mesh processing, or parallel reconstruction pipelines will be essential to enable scalable policy training.
Currently, our evaluation is restricted to exact shape matching of every cut fragment. In future work, we could introduce more generalized tasks, such as sculpting, where the objective is to match only the shape of the largest piece, thereby better probing the adaptability of our manipulation policies.
Moreover, due to the inherent fragility and resolution constraints of the MPM algorithm used in our simulation, the platform cannot fully support dexterous cutting control. Consequently, simulating highly realistic scenarios, such as objects with high stiffness like potatoes, remains challenging. Future improvements in MPM robustness and resolution will be necessary to better simulate such nuanced, realistic manipulation scenarios.

\acknowledgments{
We thank Kier Storey and Michelle Lu (former NVIDIA) for helpful discussions on soft object algorithm and implementation; Tobias W. (NVIDIA) for support with the MPM implementation in PhysX; our former colleagues on the NVIDIA PhysX team for assistance; the Vector Institute for computational resources; and the PairLab members at Georgia Tech for helpful feedback and peer review.
}

\bibliography{example}  

\newpage
\section{Appendix}

\begin{figure}[ht]
  \centering
  \includegraphics[width=0.9\linewidth]{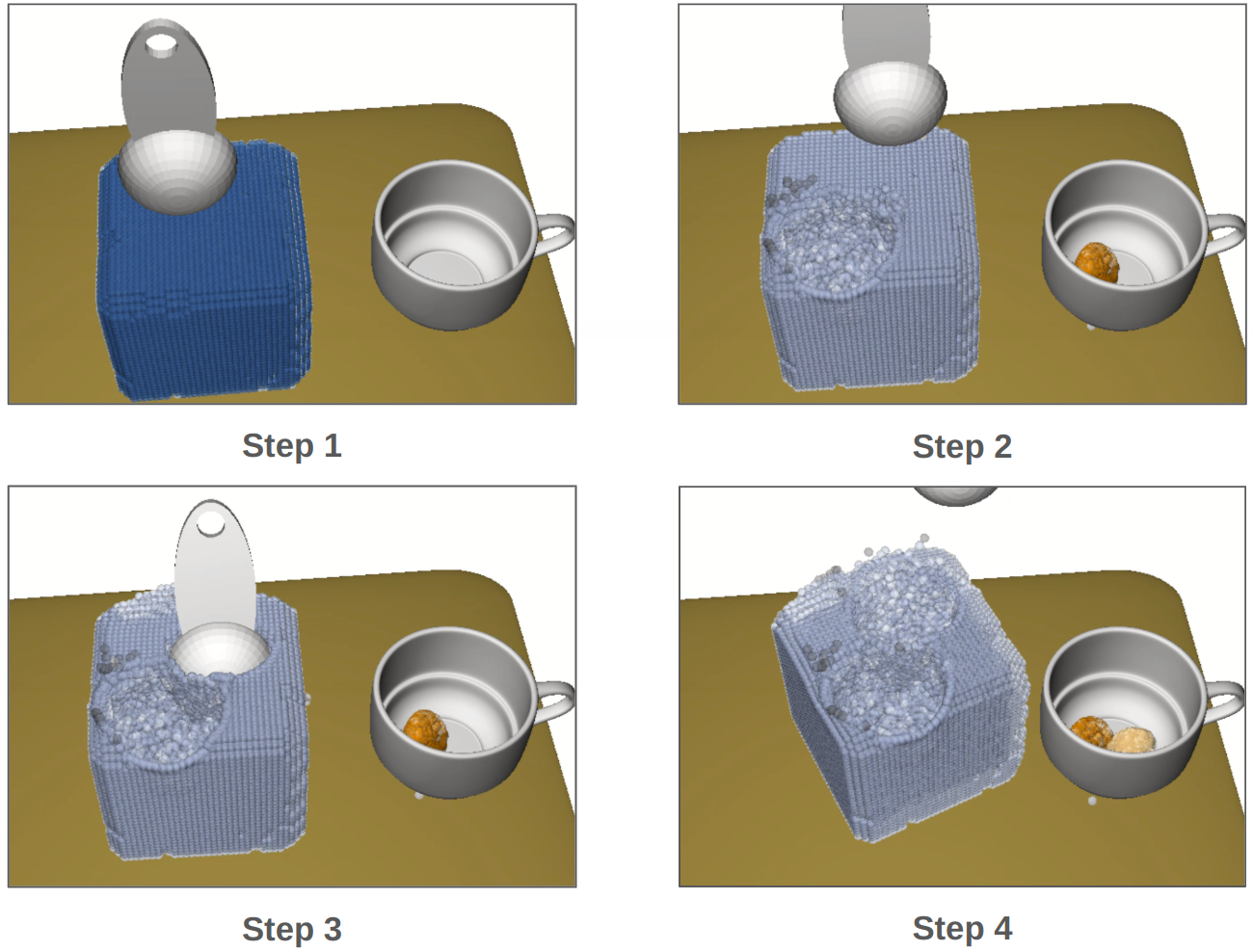}
  \caption{Ice–Cream Scooping Task. (Step 1) Initial setup of the scene, with the spoon positioned above the ice cream cube. A white cup is placed beside the ice cream block to receive the scooped pieces. (Step 2) After the first scoop, a distinct ice cream piece appears in the cup, shown in a different color to indicate it is a separate connectivity cluster. (Step 3) The spoon performs a second scooping, inserting into the ice cream cube to gather another piece. (Step 4) The task is completed with two distinct ice cream pieces, each in its own cluster, now present in the cup.}
  \label{fig:icecream_scooping}
\end{figure}

\begin{figure}[ht]
  \centering
  \includegraphics[width=0.9\linewidth]{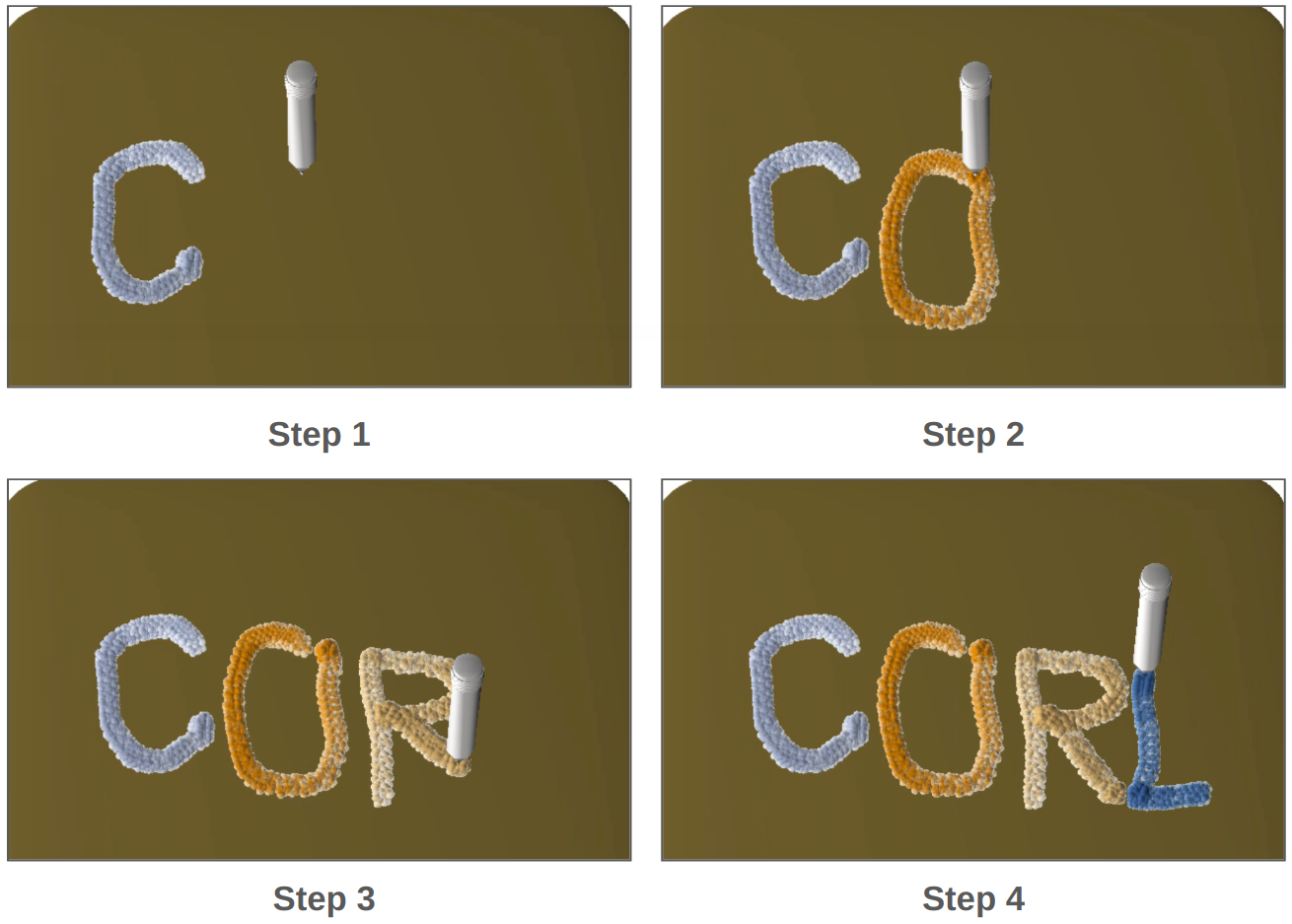}
  \caption{Cream Writing Task. (Step 1) The cream pen extrudes a continuous stroke to form the letter “C.” (Step 2) The pen lifts, repositions, and extrudes the letter “O.” (Step 3) Next, it draws “R.” (Step 4) Finally, it writes “L.” Each letter is rendered in a different color to indicate that they belong to separate connectivity clusters. The color differentiation is not manually controlled but automatically generated based on the clustering, which is leveraged during the spectral-based reward computation.}
  \label{fig:cream_writing}
\end{figure}

\subsection{Additional Manipulation Tasks}

While our main focus is on multi-step cutting, our simulator, topology discovery mechanism, and spectral reward function extend naturally to other deformable-object tasks. Below, we describe two illustrative examples that highlight the potential for extending our framework, using enumerated steps for each task’s execution.

\subsubsection{Cream Writing}
In this task, the objective is to “write” a target word (e.g., \texttt{CORL}) by extruding a soft, von Mises-plastic “cream” material onto a flat surface. The cream is modeled in MPM with low yield stress and high plasticity, allowing it to retain its shape after extrusion. The agent controls a “cream pen” at a fixed height above the table. Its state is represented by its $(x, y)$ position, and the action is a 3-dimensional vector $(\Delta x, \Delta y, b)$, where $(\Delta x, \Delta y)$ specifies the pen movement and $b \in \{0, 1\}$ toggles the extrusion. 

During the writing process, each continuous stroke formed while $b = 1$ is considered a single connectivity cluster. If the pen is lifted and re-positioned, the next stroke is assigned to a new cluster. Thus, the clustering mechanism is naturally defined by the writing process, not by explicit color control. The visualization in Figure~\ref{fig:cream_writing} uses different colors to indicate separate clusters, a distinction that becomes critical when computing the spectral-based reward.

Execution steps:
\begin{enumerate}
  \item Position the pen at the starting point for the letter “C,” toggle $b = 1$, and move rightward to trace the stroke, forming the first connectivity cluster.
  \item Lift the pen, move to the position for the letter “O,” toggle $b = 1$, and extrude to form the second cluster.
  \item Reposition to the third slot, toggle $b = 1$, and draw the letter “R,” creating a new cluster.
  \item Finally, move to the fourth slot, toggle $b = 1$, and write the letter “L,” forming the final cluster.
\end{enumerate}

Our spectral reward function evaluates how well each cluster corresponds to the intended letter shape. This clustering mechanism provides a natural way to segment distinct components, allowing the reward function to assess both spatial structure and connectivity. However, the sequential planning of strokes remains a complex decision-making challenge, distinct from the cutting task, and thus extending PDDP to handle such “additive” tasks is a promising direction for future work.

\subsubsection{Ice-Cream Scooping}
In this task, the objective is to scoop a chunk from a block of “pudding”-like ice cream (elastoplastic MPM) and deposit it into a cup. The agent controls a spoon at a fixed height, using a 2D $(x, y)$ action space to position the scoop. Once positioned, the agent executes a predefined scooping primitive as follows:

Execution steps:
\begin{enumerate}
  \item Lower the spoon into the ice cream block to initiate contact.
  \item Translate the spoon forward to penetrate and gather the material.
  \item Lift the spoon while rotating it to maintain a horizontal orientation, preventing spillage.
  \item Move the spoon above the cup.
  \item Tilt or open the spoon to release the scooped chunk into the cup.
\end{enumerate}

The scooped chunk forms a distinct connectivity cluster, separate from the remaining block. As illustrated in Figure~\ref{fig:icecream_scooping}, each scooped piece is automatically assigned a different color to indicate it belongs to a separate cluster. This distinction is crucial when computing our spectral-based reward, as it allows for quantifying successful segmentation and accurate placement of the scooped pieces.

\subsubsection{Discussion}
These two examples demonstrate the extensibility of our MPM-based environment, topology discovery, and spectral reward function to a broader range of deformable manipulation tasks beyond multi-step cutting. The use of distinct clusters to represent separate connectivity components is a key mechanism for computing spectral-based rewards, applicable in both “additive” and “scooping” tasks. Designing specialized action primitives and learning-based planners for these scenarios remains a promising direction for future exploration.

\subsection{Pyramid Cutting Task}

\begin{figure}[ht]
  \centering
  \includegraphics[width=0.95\linewidth]{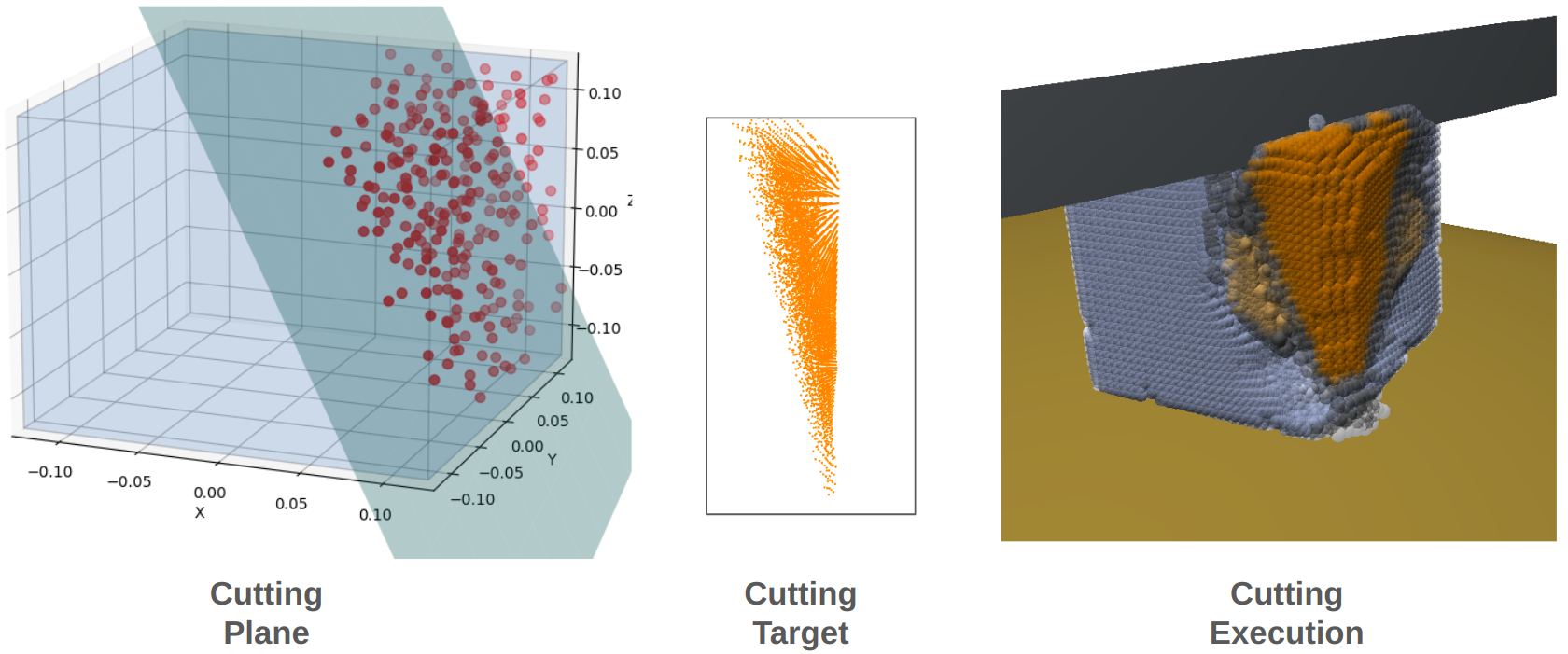}
  \caption{Pyramid Cutting Task. (Left) Cutting Plane: Visualization of the cutting plane planned by the MPPI controller based on our spectral reward function. The plane is strategically positioned to carve out a pyramid-shaped fragment from the corner of the cube. (Middle) Cutting Target: The target goal shape, shown as a point cloud representation, highlighting the desired pyramid structure that serves as a guide for the cutting task. (Right) Cutting Execution: The scene depicts the execution of the planned cutting trajectory, where the knife removes the corner section to achieve the specified pyramid shape. The orange particles indicate the material identified as the target segment to be removed, aligning with the planned cutting plane.}
  \label{fig:pyramid_cut}
\end{figure}

\paragraph{Motivation and Objective.} 
To further evaluate the generalization capability of our spectral reward function to more complex and arbitrary goal shapes, we introduce the pyramid cutting task (Figure~\ref{fig:pyramid_cut}). In contrast to the predefined slicing, sticking, and dicing tasks, this experiment requires the agent to carve out a corner segment from a cubic block to produce a pyramid-shaped fragment. The target shape is defined as a triangular pyramid with sloped surfaces, representing a more intricate and asymmetrical geometry than previous tasks.

The objective of this experiment is to assess whether the MPPI planning method, driven solely by our spectral reward function, can effectively discover a plausible cutting trajectory to achieve the desired pyramid shape without any task-specific tuning or retraining. This setup challenges the reward function to guide cutting actions that align with more complex, multi-faceted surfaces, testing its robustness and generalizability.

\paragraph{Takeaway.} 
The results of the pyramid cutting task clearly demonstrate that our spectral reward function effectively generalizes to more intricate geometric goals, even in cases involving angled and asymmetric surfaces. As shown in Figure~\ref{fig:pyramid_cut}, the MPPI planner successfully identifies a cutting plane that produces a distinct pyramid-shaped fragment from the cube’s corner, aligning well with the target structure. The qualitative alignment between the planned cutting plane and the final extracted segment validates that the reward function remains a reliable guidance signal, even for complex and non-standard shapes. This outcome highlights the adaptability of our framework, suggesting its potential to handle more sophisticated and arbitrary cutting tasks in future applications.

\subsection{Perception Model Architecture Details}

In practice, we design two distinct styles of perception models to explore their effectiveness in our setup: a graph-based model and a point-based model. The graph-based model leverages graph convolutional networks (GCNs) and joint attention mechanisms to process structured graph data, while the point-based model employs a hierarchical PointNet-style architecture using SPALPA blocks to handle point cloud data. After extensive evaluation, we adopt the graph-based model as our final choice. This decision is driven by the significant reduction in parameter count (approximately four times fewer parameters) and faster training and inference speed compared to the point-based model. The PointNet-based architecture, though capable of capturing finer local features, incurs substantial computational overhead due to its deeper SPALPA structure and multi-scale processing layers.

\subsubsection{PointNet-Based Perception Model Architecture}

The PointNet-based architecture processes point cloud data through hierarchical SPALPA blocks for multi-scale feature extraction. The key components are:

\begin{itemize}
    \item \textbf{Topological Encoder:} 
    The primary perception encoder in this architecture, responsible for extracting features from 32-dimensional topological inputs. It comprises 5 SPALPA blocks that progressively increase feature dimensions as follows:
    \[
    96 \rightarrow 192 \rightarrow 384 \rightarrow 768 \rightarrow 1536
    \]

    Each SPALPA block includes:
    \begin{itemize}
        \item \textbf{Local Attention:} Extracts localized spatial relationships through `Conv2d`.
        \item \textbf{Global Attention:} Aggregates broader context using multi-scale convolutional layers.
        \item \textbf{Grouper:} Implements `QueryAndGroup` for spatial neighborhood aggregation.
    \end{itemize}

    \item \textbf{Action Encoder:}
    Encodes 2-dimensional action inputs using the same hierarchical structure as the Topological Encoder, ensuring feature alignment and consistency.

    \item \textbf{Cross-Attention Network:}
    Integrates topological and action features through CrossAttnNet modules, employing PreNorm Attention, cross-attention, and gated residual connections.

    \item \textbf{Decoder:}
    Reconstructs the feature maps and propagates them to the original point cloud resolution using `FeaturePropagation` blocks, merging multi-scale features.

    \item \textbf{Output Heads:}
    \begin{itemize}
        \item \textbf{Segmentation Head:} Projects features to segmentation classes using `Conv1d` layers.
        \item \textbf{Point Cloud Head:} Predicts point coordinates through residual blocks and a final `Linear` layer.
    \end{itemize}

    Despite its comprehensive feature extraction capability, the PointNet-based model incurs considerable computational overhead, resulting in slower training and inference compared to the graph-based model.
\end{itemize}

\subsubsection{Graph Network Architecture}

The Graph Network architecture processes structured graph data using GCN layers and attention-based mechanisms. The core perception module in this architecture is the \textbf{t\_graph\_encoder}, which encodes topological information through GCN layers.

\begin{itemize}
    \item \textbf{t\_graph\_encoder:}
    The designated perception module in this architecture, processing node and label embeddings through GCN layers structured as:
    \[
    3 \rightarrow 96 \rightarrow 96
    \]
    Key submodules include:
    \begin{itemize}
        \item \textbf{Node Encoder:} Projects 3-dimensional node features to a 96-dimensional latent space.
        \item \textbf{Label Encoder:} Processes 32-dimensional node labels through GCN layers.
        \item \textbf{Embedding Encoder:} Integrates 192-dimensional precomputed embeddings into a unified 96-dimensional space.
    \end{itemize}

    \item \textbf{a\_graph\_encoder:}
    Processes action-related graph data using a similar structure to the `t\_graph\_encoder`, but with 2-dimensional action inputs.

    \item \textbf{Joint Graph Transformer:}
    Integrates features from both graph encoders through multi-layer attention and cross-attention modules:
    \[
    96 \rightarrow 64 \rightarrow 32 \rightarrow 16 \rightarrow 8 \rightarrow 4
    \]

    \item \textbf{Query Point Encoder:}
    Processes query nodes in the graph structure through `Linear` and `ReLU` layers:
    \[
    3 \rightarrow 64 \rightarrow 96
    \]

    \item \textbf{Output Heads:}
    \begin{itemize}
        \item \textbf{Node Position Head:} Outputs node coordinates through residual blocks.
        \item \textbf{Node Feature Head:} Projects node features to a 32-dimensional space.
    \end{itemize}

    The graph-based model provides a more parameter-efficient and computationally feasible architecture while maintaining robust feature extraction and representation capabilities, making it the preferred choice in our implementation.
\end{itemize}

\paragraph{Summary:}
We explore both PointNet-based and Graph Network architectures to identify the optimal perception model for our setup. While the PointNet-based model employs extensive SPALPA blocks for multi-scale feature extraction, its computational cost and parameter count are substantially higher than the graph-based model. Consequently, we adopt the graph-based architecture, leveraging the \textbf{t\_graph\_encoder} as the primary perception module due to its significantly lower parameter count, faster training and inference speeds, and effective integration of topological and action features.

\subsubsection{Empirical Evaluation of Perception Models}

\begin{table*}[t] \LARGE
  \centering
  \caption{\small
  \textbf{Empirical Studies of Pretraining Strategies for Perception Models:}
  We evaluate the effectiveness of graph-based and point-based perception models under various pretraining strategies, including sorting the input point cloud, applying Gaussian noise for data augmentation, and downsampling the input points before processing. The final model selected for deployment is the Graph-Based model with sorting, Gaussian noise, and 256 downsampled points, balancing accuracy and computational efficiency.
  }
  \label{table:pretrain_eval}
  \renewcommand{\arraystretch}{1.5}
  \setlength{\tabcolsep}{15pt}
  \resizebox{0.95\linewidth}{!} {
  \begin{tabular}{l||ccc|cc}
  \toprule
  \rowcolor[HTML]{FFDEB4}
  \textbf{Perception Model} & \textbf{With Sort} & \textbf{Gaussian Noise} & \textbf{Downsample Points} & \textbf{Seen Accuracy} & \textbf{Unseen Accuracy} \\
  \midrule

  \rowcolor[HTML]{FFFFFF}
  Graph-Based & \ding{55} & \ding{55} & 128 & 66.4 & 48.7 \\\cmidrule{1-6}

  \rowcolor[HTML]{EFEFEF}
  Graph-Based & \checkmark & \ding{55} & 128 & 73.8 & 52.1 \\\cmidrule{1-6}

  \rowcolor[HTML]{FFFFFF}
  Graph-Based & \checkmark & \checkmark & 128 & 75.4 & 69.3 \\\cmidrule{1-6}

  \rowcolor[HTML]{EFEFEF}
  \textbf{Graph-Based (Final)} & \textbf{\checkmark} & \textbf{\checkmark} & \textbf{256} & \textbf{77.1} & \textbf{74.1} \\\cmidrule{1-6}

  \rowcolor[HTML]{FFFFFF}
  PointNet-Based & \ding{55} & \checkmark & 2048 & 73.2 & 61.2 \\\cmidrule{1-6}

  \rowcolor[HTML]{EFEFEF}
  PointNet-Based & \checkmark & \checkmark & 512 & 79.2 & 75.4 \\

  \bottomrule
  \end{tabular}
  }
\end{table*}

\paragraph{Discussion and Model Selection:}
Table~\ref{table:pretrain_eval} presents a comparative evaluation of graph-based and point-based perception models under various pretraining strategies. The primary factors analyzed include sorting based on the x-axis, Gaussian noise for data augmentation, and the degree of point cloud downsampling.

\begin{itemize}
    \item \textbf{Sorting and Gaussian Noise:} Incorporating sorting and Gaussian noise consistently improves model accuracy for both seen and unseen data. For the graph-based model, enabling both sorting and Gaussian noise increased unseen accuracy from 48.7\% to 69.3\% under a 128-point downsampling setting.

    \item \textbf{Downsampling Strategy:} The impact of point cloud resolution is evident as increasing the number of points from 128 to 256 in the graph-based model further improved unseen accuracy to 74.1\%. This result underscores the importance of maintaining sufficient point density to preserve critical spatial information.

    \item \textbf{Comparison of Models:} The point-based model with 512 points achieves 75.4\% unseen accuracy, slightly outperforming the graph-based model with 256 points. However, the computational overhead associated with the point-based model is significantly higher, consuming approximately four times more parameters and resulting in slower training and inference.

\end{itemize}

\paragraph{Final Model Selection:}
Based on the empirical findings, the graph-based model with sorting, Gaussian noise, and 256 downsampled points is selected as the final perception model for deployment. This configuration achieves a balanced trade-off between computational efficiency and segmentation accuracy, making it a practical choice for large-scale robotic manipulation tasks.

\subsection{Policy Model Architecture Details}

In our framework, the policy model architecture is structured to handle skill-conditioned action generation through a structured diffusion process. The model, named \textbf{PDDP (Particle-Based Diffusion Policy)}, leverages adaptive normalization, timestep embeddings, and modular diffusion blocks to effectively model action trajectories. Notably, the observation encoder in PDDP utilizes the perception encoder previously defined in the \textbf{Perception Model Architecture} section, ensuring consistent feature extraction and representation across the pipeline.

\subsubsection{Policy Model Architecture - PDDP}

The PDDP architecture is structured as follows:

\paragraph{1. Encoders:}

\begin{itemize}
    \item \textbf{Observation Encoder:}
    The observation encoder adopts the perception encoder design outlined previously, specifically utilizing the graph-based \textbf{t\_graph\_encoder} as the primary perception module. This encoder processes topological node features through a GCN and includes three key submodules:
    \begin{itemize}
        \item \textbf{Node Encoder:} Projects 3-dimensional node features to a 96-dimensional latent space through a 2-layer GCN:
        \[
        3 \rightarrow 96 \rightarrow 96
        \]
        \item \textbf{Label Encoder:} Transforms 32-dimensional node labels to a 96-dimensional space using the same GCN structure.
        \item \textbf{Embedding Encoder:} Integrates 192-dimensional precomputed embeddings, projecting them to 96 dimensions.
    \end{itemize}
    The outputs from each encoder component are processed through a shared `LayerNorm` to stabilize feature representation.

    \item \textbf{Goal Encoder:}
    Encodes 3-dimensional goal vectors through a fully connected network:
    \[
    3 \rightarrow 64 \rightarrow 96
    \]
    ReLU activation is applied to the intermediate layer.

    \item \textbf{Timestep Embedding (sigma\_map):}
    Encodes the diffusion timestep using a 2-layer MLP:
    \[
    256 \rightarrow 128 \rightarrow 128
    \]
    A SiLU activation is applied after the first linear layer to maintain smooth gradient flow.

    \item \textbf{Action History Layer:}
    Integrates action history features through a fully connected layer:
    \[
    21 \rightarrow 128
    \]

    \item \textbf{Conditional Layer:}
    Aggregates timestep, goal, and action history embeddings into a unified feature space:
    \[
    256 \rightarrow 128
    \]
\end{itemize}

\paragraph{2. Backbone - Diffusion Blocks (DDiTBlock):}

The core processing backbone of PDDP consists of a sequence of 12 \textbf{DDiTBlocks}. Each DDiTBlock contains:

\begin{itemize}
    \item \textbf{Self-Attention Layer:} 
    Applies self-attention to the feature space with 96-dimensional query, key, and value projections:
    \[
    96 \rightarrow 288 \rightarrow 96
    \]
    This allows the model to capture dependencies across all points in the point cloud, facilitating information exchange across nodes.

    \item \textbf{MLP Layer:} 
    Implements a 2-layer MLP with GELU activation:
    \[
    96 \rightarrow 384 \rightarrow 96
    \]
    This structure refines the feature representation after the attention operation, maintaining non-linear feature transformations.

    \item \textbf{Dropout:} 
    Applied to both the attention and MLP layers with a probability of 0.4 to mitigate overfitting and stabilize training.

    \item \textbf{Adaptive LayerNorm (adaLN):}
    Adaptive LayerNorm is applied to each block using modulation inputs derived from both the timestep embedding and goal encoder. The modulation layer is structured as:
    \[
    128 \rightarrow 576
    \]
    This mechanism enables each DDiTBlock to dynamically adjust its feature scaling based on task-specific conditions.
\end{itemize}

\paragraph{3. Output Layer:}

The final output layer processes the feature representation generated by the diffusion blocks to predict binary classification logits for each point in the point cloud. The output shape is defined as:

\[
\text{Output Shape: } [\text{batch size}, \text{num points}, 2]
\]

A linear projection is applied to transform the 96-dimensional feature space into 2-dimensional logits, representing the binary classification output for each point. This design aligns the output structure with the expected action space while maintaining consistency across the diffusion layers.

\paragraph{Summary:}
The PDDP architecture is structured to leverage a modular pipeline consisting of encoders for observation, goal, and timestep embeddings, a backbone of 12 DDiTBlocks with adaptive normalization and self-attention mechanisms, and a binary classification output layer for per-point action prediction. The use of the \textbf{t\_graph\_encoder} as the primary perception module ensures consistent feature extraction and integration across both the perception and policy networks, promoting robust action generation in diverse manipulation tasks.

\subsection{Material Point Method in Details}\label{app:mpm}

The Material Point Method (MPM) is a hybrid Lagrangian–Eulerian scheme originally introduced by Sulsky \emph{et al.} for solid mechanics.  In MPM, state variables (mass, momentum, deformation) are carried on material points (particles), while a fixed Eulerian grid is used to solve the equations of motion.  This split representation combines the mesh‐free advantages of particle methods with the stability and boundary‐handling capabilities of grid methods.

\paragraph{Governing Equations}  
MPM begins from the continuum balance laws in Eulerian form:
\[
\frac{D\rho}{Dt} + \rho\,\nabla\!\cdot v = 0,
\qquad
\rho\,\frac{Dv}{Dt} = \nabla\!\cdot\sigma + \rho\,g,
\]
where $\rho$ is the mass density, $v$ the velocity field, $\sigma$ the Cauchy stress tensor, and $g$ the body‐force (e.g.\ gravity).  The material derivative $D/Dt$ captures convection of field quantities with the flow.

\paragraph{Weak Form and Discretization}  
To derive a tractable discretization, one multiplies the momentum equation by a test function $q(x)$ and integrates over the domain $\Omega^n$ at time step $n$.  Integration by parts moves spatial derivatives onto $q$, yielding the weak form.  We then partition $\Omega^n$ into particle subdomains $\Omega_p^n$, and approximate both trial and test functions using B‐spline shape functions $N_i(x)$ centered at grid nodes $i$.  This Galerkin‐style projection leads to discrete nodal equations that are assembled via sums over particles.

\paragraph{Particle–to–Grid (P2G) Transfer}  
Each particle $p$ holds:
\[
\{\,m_p,\;V_p,\;v_p,\;F_p\}
\]
for mass $m_p$, volume $V_p$, velocity $v_p$, and deformation gradient $F_p$.  To project onto the grid, we compute at each node $i$:
\[
m_i \;=\;\sum_{p} m_p\,N_i(x_p),
\quad
(mv)_i \;=\;\sum_{p} m_p\,v_p\,N_i(x_p),
\quad
f^{\rm int}_i \;=\;- \sum_{p} V_p\,\sigma_p\,\nabla N_i(x_p),
\]
where $\sigma_p$ is obtained from the chosen constitutive law (e.g.\ hyperelastic) evaluated at $F_p$.  These transfers ensure exact conservation of mass and momentum.

\paragraph{Grid Update}  
On the Eulerian grid, we update nodal momentum via a symplectic (explicit) Euler step:
\[
(mv)_i^{n+1}
= (mv)_i^{n} + \Delta t\bigl(f_i^{\rm int} + m_i\,g\bigr),
\qquad
v_i^{n+1} = \frac{(mv)_i^{n+1}}{m_i}.
\]
This step advances velocities under both internal stresses and external forces, while preserving stability for moderate time steps.

\paragraph{Grid–to–Particle (G2P) Transfer}  
After updating the grid, we interpolate back to particles:
\[
v_p^{n+1}
= \sum_{i} N_i(x_p)\,v_i^{n+1},
\qquad
C_p
= \sum_{i} v_i^{n+1}\,\nabla N_i(x_p)^{T},
\]
where $C_p$ is an affine velocity gradient matrix (used in variants like APIC) that captures sub‐cell velocity variation.

\paragraph{Deformation Gradient Update}  
The particle deformation gradient evolves according to
\[
\frac{dF}{dt} = (\nabla v)\,F.
\]
Using the interpolated velocity gradient $C_p$, we discretize:
\[
F_p^{\,n+1}
= \bigl(I + \Delta t\,C_p\bigr)\,F_p^{\,n},
\]
so that $F_p$ accumulates the local deformation history on each particle.

\paragraph{Stress Computation}  
For hyperelastic materials, one defines a strain‐energy density $\Psi(F)$ and computes the first Piola–Kirchhoff stress
\[
P_p = \frac{\partial \Psi}{\partial F}\bigl(F_p\bigr),
\]
then the corresponding Cauchy stress
\[
\sigma_p = \frac{1}{\det F_p}\,P_p\,F_p^{T}.
\]
This stress is used in the P2G transfer to produce internal forces.

\paragraph{Algorithm Summary}  
At each timestep, MPM executes:
\begin{enumerate}
  \item \textbf{P2G:} Transfer $\{m_p,\;v_p,\;F_p\}$ to grid nodes $\{m_i,\;v_i,\;f_i\}$.  
  \item \textbf{Grid Update:} Integrate nodal momentum, compute $v_i^{n+1}$.  
  \item \textbf{G2P:} Interpolate $v_i^{n+1}$ and $\nabla v$ back to particles.  
  \item \textbf{State Update:} Update each $F_p^{\,n+1}$ and compute $\sigma_p$.  
  \item \textbf{Advection:} Move particles: $x_p^{n+1} = x_p^n + \Delta t\,v_p^{n+1}$.  
  \item \textbf{Reset:} Clear grid variables for the next iteration.
\end{enumerate}

Thanks to its hybrid nature, MPM can robustly simulate extreme deformations, fracture propagation, and multi‐body contact without remeshing, making it a powerful tool in graphics, engineering, and robotic manipulation contexts.

\subsection{Particle-based Damage-tracking and Topological Reconstruction in Details}\label{app:particle_damage} 

\begin{figure}[t]
  \centering
  \includegraphics[width=0.6\linewidth]{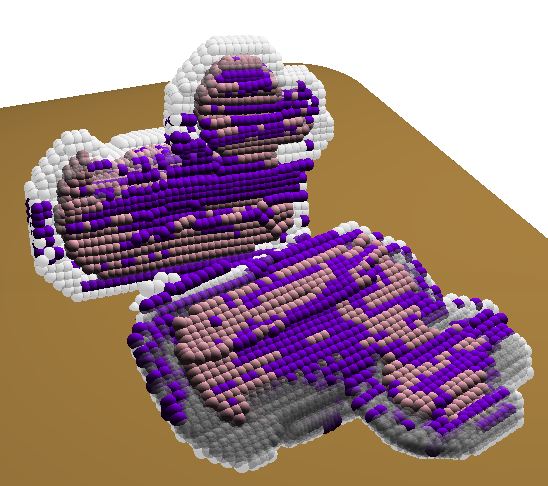}
  \caption{Particle-based damage-tracking and topological reconstruction during a cutting action. Gray and pink particles remain intact, while damaged particles (purple) mark the cut interface. From these damage signals we build an implicit SDF, extract the zero-isosurface via Marching Cubes, and then cluster connected components to recover discrete object segments.}
  \label{fig:particle_topology}
\end{figure}

We propose a robust particle-based method to accurately determine whether a cutting action successfully separates an object into discrete pieces. Our approach integrates particle-level damage tracking with topological surface reconstruction using the Material Point Method (MPM) as shown in Figure~\ref{fig:particle_topology}. Within our MPM framework, each particle's deformation state is computed, and we define a particle as ``damaged'' based on critical compression and stretch thresholds in its deformation gradient \(\mathbf{F}\). Formally, a particle \(p\) is classified as damaged if:
\begin{equation}
J_p \leq (1 - \epsilon_c)^m \quad\text{or}\quad J_p \geq (1 + \epsilon_s)^m,
\label{eq:damage_particle}
\end{equation}
where \(J_p = \det(\mathbf{F}_p)\), \(\epsilon_c, \epsilon_s\) denote critical compression and stretch values, and \(m\) is a material-specific sensitivity exponent.

In practice, we select the compression and stretch thresholds \(\epsilon_c\) and \(\epsilon_s\) so that they correspond directly to the maximum volumetric change a particle can undergo before being flagged as damaged. Recall from Eq.~\eqref{eq:damage_particle} that
\[
J_p = \det(\mathbf{F}_p)
\]
measures the local volume ratio of particle~\(p\).  For example, setting
\[
\epsilon_c = 0.025,\qquad \epsilon_s = 0.01
\]
means that a particle whose volume has decreased by more than \(2.5\%\) (\(J_p \le (1 - 0.025)^m\)) or increased by more than \(1\%\) (\(J_p \ge (1 + 0.01)^m\)) is classified as damaged.  These values are chosen to be small enough to detect the onset of fracture in brittle materials, yet large enough to avoid false positives under normal elastic deformation.

The exponent \(m\) modulates how sharply damage accumulates once \(J_p\) deviates from unity: a larger \(m\) yields an abrupt transition from “healthy” to “damaged,” whereas a smaller \(m\) produces a more gradual damage accumulation.  We calibrate \(\epsilon_c\), \(\epsilon_s\), and \(m\) using simulated uniaxial compression and tension tests: a small block is deformed at a constant strain rate, we record its volumetric Jacobian history, and then choose the smallest thresholds that cleanly separate reversible (elastic) behavior from irreversible damage.

For more ductile or plastic materials, one might increase the stretch threshold (e.g.\ \(\epsilon_s \approx 0.05\)) and use a lower sensitivity exponent (e.g.\ \(m=1\) or \(2\)) to model gradual yielding.  Conversely, for near-brittle media a tighter compression bound (e.g.\ \(\epsilon_c \approx 0.01\)) and higher exponent (e.g.\ \(m\ge4\)) better capture sudden fracture.  By anchoring these parameters to physically measurable strain limits, our framework remains both interpretable and readily tunable across a wide range of material behaviors.

For more complex yielding materials, such as those modeled with von Mises plasticity, damage occurs when the yielding stress \(\sigma_y\) falls below zero after particle softening:
\begin{equation}
\sigma_y^{(t+1)} = \sigma_y^{(t)} - \gamma \|\Delta\epsilon_p\|,\quad\text{with}\quad \text{Damaged}_p =
\begin{cases}
1 & \text{if }\sigma_y^{(t+1)} \leq 0,\\[2pt]
0 & \text{otherwise,}
\end{cases}
\label{eq:vonmises_damage}
\end{equation}
where \(\gamma\) is the particle softening coefficient and \(\Delta\epsilon_p\) represents incremental plastic strain.

Once particles become damaged, their mechanical properties are altered irreversibly, enabling consistent tracking of the damaged state throughout subsequent simulation steps.

To reconstruct and evaluate object topology following a cut, we perform explicit topological segmentation leveraging the particle damage signals and knife trajectory data. We track the knife trajectory precisely during interactions, ensuring a clean spatial separation between object segments. After the completion of a cut action, defined by a stationary knife and changed particle damage states, we reconstruct the object's surface mesh from particle data through an implicit surface representation using Signed Distance Fields (SDFs) and the Marching Cubes algorithm.

Formally, given the set of particles \(\mathcal{P}\) and knife trajectory \(\mathcal{T}\), we first construct an implicit SDF representation \(SDF(\mathbf{x})\) of the object's spatial domain:
\begin{equation}
SDF(\mathbf{x}) = \min_{p \in \mathcal{P}} \|\mathbf{x}-\mathbf{x}_p\| - r_p,
\end{equation}
where \(\mathbf{x}_p\) is the position of particle \(p\), and \(r_p\) is its effective influence radius. Critically, we incorporate the knife trajectory \(\mathcal{T}\) by explicitly marking trajectory regions within the SDF, enforcing a strict gap devoid of particles. This guarantees that no particle fills the knife's trajectory space, creating a clear geometric boundary between separated pieces.

We perform surface reconstruction by extracting the zero-isosurface of the computed SDF using a GPU-accelerated Marching Cubes implementation provided by Warp~\cite{nvidia2022warp}. Specifically, we first define a discrete volumetric grid around particle positions, evaluate the SDF values, and then generate vertices and faces for the surface mesh:
\begin{equation}
\text{vertices, faces} = \text{MarchingCubes}(SDF(\mathbf{x})=0).
\end{equation}

This mesh reconstruction process ensures explicit representation of separated object components. After extracting mesh segments, we apply Laplacian smoothing to refine surface quality:
\begin{equation}
\mathbf{v}_i \leftarrow \mathbf{v}_i + \alpha \sum_{j\in \mathcal{N}(i)} (\mathbf{v}_j - \mathbf{v}_i),
\end{equation}
where \(\mathbf{v}_i\) is a vertex in the mesh, \(\mathcal{N}(i)\) are neighboring vertices, and \(\alpha\) is a smoothing factor.

Subsequently, we cluster these reconstructed mesh segments into distinct topological pieces by performing connected component analysis on the extracted mesh faces. Each particle is then assigned a cluster identity by evaluating its spatial relationship to these mesh segments. Given particle positions \(\mathbf{x}_p\) and cluster meshes, we perform the following SDF-based assignment:
\begin{equation}
\text{Cluster}_p =
\begin{cases}
i & \text{if } SDF_i(\mathbf{x}_p) < \tau,\\[2pt]
-1 & \text{otherwise,}
\end{cases}
\end{equation}
where \(SDF_i\) is the SDF for cluster mesh \(i\), and \(\tau\) is a proximity threshold.

After assigning new cluster identities, we update the global object topology by comparing current particle clusters to previously stored cluster identifiers. New cluster IDs are assigned whenever new separate segments emerge, thus preserving consistent object segmentation across simulation timesteps.

This methodical combination of particle damage tracking, implicit surface reconstruction, and mesh-based clustering provides a robust and efficient solution for detecting and managing the topological changes that occur during robotic cutting tasks. Our approach ensures accurate evaluation of cutting outcomes essential for effective policy learning and performance assessment.

\subsection{Pose-Invariant Shape Evaluation via Spectral Analysis in Details}\label{app:pose_invar} 

After discovering object topology, we require a robust metric to evaluate how closely the segmented fragments match the intended goal shape. Traditional point-set distances, such as Chamfer or Earth Mover’s Distance, are sensitive to pose variations and require explicit alignment. To overcome these limitations, we introduce a spectral-based reward derived from the intrinsic geometry of shapes, inherently invariant to rigid transformations.

Given a point cloud \(X = \{x_i\}_{i=1}^n\), we construct a \(k\)-nearest neighbor graph where the edge weight between points \(i\) and \(j\) is defined as
\begin{equation}
W_{ij} = \exp\left( -\frac{d_{ij}^2}{\sigma^2} \right),
\end{equation}
with \(d_{ij}\) the geodesic distance between points and \(\sigma\) a scaling parameter. We then build the degree matrix \(D\) and the combinatorial Laplacian:
\begin{equation}
D_{ii} = \sum_j W_{ij}, \quad L = D - W.
\end{equation}
Eigen-decomposition of \(L\) yields the smallest \(k\) eigenpairs:
\begin{equation}
L \Phi = \Lambda \Phi,
\end{equation}
where \(\Lambda = \text{diag}(\lambda_1, \dots, \lambda_k)\) contains eigenvalues and \(\Phi = [\phi_1, \dots, \phi_k]\) are the corresponding eigenvectors. These spectral descriptors capture intrinsic shape information invariant to isometries.

Given two shapes \(X\) and \(Y\), we define their spectral distance as:
\begin{equation}
d_{\text{spec}}(X, Y) = \alpha \|\Lambda_X - \Lambda_Y\|_2^2 + \beta \|\Phi_X^\top \Phi_X - \Phi_Y^\top \Phi_Y\|_F^2,
\end{equation}
where \(\alpha, \beta\) balance the contributions of eigenvalue and eigenvector differences.

To transform spectral distance into a reward, we apply a piecewise linear mapping with a critical threshold \(\tau\):
\begin{equation}
R(d) =
\begin{cases}
R_{\text{max}} - \gamma d, & d \leq \tau, \\\\
R_{\text{max}} - \gamma \tau - \delta (d - \tau), & d > \tau,
\end{cases}
\end{equation}
where \(\gamma\) and \(\delta\) are decay rates (\(\delta > \gamma\)), ensuring a gradual reward decline for small errors and sharper penalties for large deviations.

For objects segmented into multiple fragments, we compute the spectral distance and associated reward for each fragment pair individually, then accumulate the total reward as:
\begin{equation}
R_{\text{total}} = \kappa \sum_i R(d_i),
\end{equation}
with \(\kappa\) a global scaling factor ensuring consistent reward magnitudes across tasks.

This formulation provides an efficient, continuous, and pose-invariant reward signal for evaluating and planning goal-conditioned robotic cutting actions.

In order to evaluate the efficient and pose-invariant property, we conduct the following experiments:

\begin{figure}[t]
  \centering
  \includegraphics[width=0.8\linewidth]{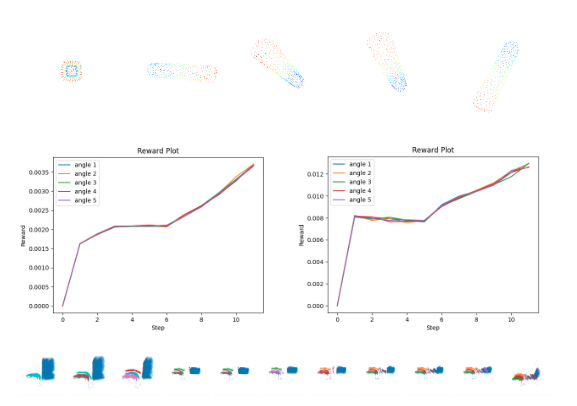}
  \caption{
    \textbf{Pose‐Invariance Evaluation of the Spectral Reward.}
    \textbf{Top:} Five ideal stick fragments (\(5\times5\times32\)) rotated through distinct angles.
    \textbf{Middle:} Reward‐versus‐cut‐step curves for each rotation, computed using dense particles (left) and mesh‐surface particles (right) with \texttt{num\_point}=512 and \(k=30\).
    \textbf{Bottom:} Corresponding fragment point‐cloud sequences along the default cutting trajectory.
    The nearly identical curves across all rotations confirm that our spectral reward is invariant to rigid transformations of the ideal shape.
  }
  \label{fig:pose_invariance}
\end{figure}

\paragraph{Parameter Definitions and Pose‐Invariance Observations}
Our spectral‐reward pipeline relies on three key configuration parameters and a fixed cutting trajectory.  Specifically:
\begin{itemize}
  \item \texttt{num\_point} (\(=512\)) determines how many points are uniformly sampled from the fragment’s point cloud to serve as vertices in the spectral graph.  A larger \texttt{num\_point} yields finer geometric resolution but increases graph‐construction cost, whereas a smaller value speeds up computation at the risk of losing subtle shape details.  
  \item \(k\) (\(=30\)) defines the neighborhood size in the \(k\)-nearest‐neighbor graph: each sampled point connects to its \(k\) closest neighbors (by Euclidean distance).  This choice balances local connectivity (capturing fine features) against spectral stability (avoiding overly dense graphs that blur intrinsic geometry).  
  \item \emph{Trajectory} refers to the fixed sequence of knife motions (the “default” cutting action) applied to all shapes.  At each discrete timestep along this trajectory, we recompute the fragment’s spectral descriptor and record the resulting reward value.  
\end{itemize}
In our experiment shown in Figure~\ref{fig:pose_invariance}, we generate five identical “stick” fragments (\(5\times5\times32\)) and rotate each by a distinct yaw angle before sampling and evaluation.  We then compute two sets of reward‐vs.‐step curves—one using all dense particles, the other using only mesh‐surface particles.  As shown in the overlaid plots, all five curves coincide almost exactly for both representations.  This confirms that, under our sampling density (\texttt{num\_point}=512) and graph connectivity (\(k=30\)), the spectral reward is effectively invariant to rigid rotations of the ideal shape.

\begin{figure}[t]
  \centering
  \includegraphics[width=0.8\linewidth]{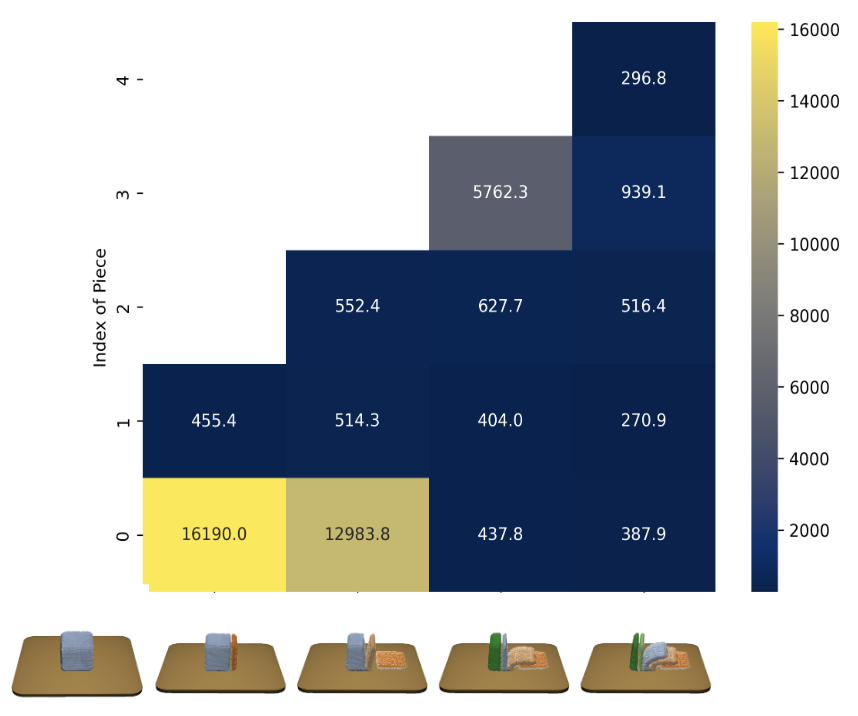}
  \caption{Pairwise spectral‐loss matrix between each cutting state (rows) and each piece within the state (columns). Lower values indicate closer intrinsic geometry matches, enabling qualitative MPPI data collection.}
  \label{fig:loss_matrix}
\end{figure}

\paragraph{Multi-step Spectral‐Loss Evaluation}
To verify that our spectral reward correctly identifies the intended fragment at each cutting step, we conducted a pairwise matching experiment on tele‐operated MPPI data.  After segmenting the object into its constituent pieces, we compute the spectral distance \(d_{\rm spec}(X_{s_i},X_{p_j})\) between the shape at cutting state \(s_i\) and each fragment \(p_j\), for \(i,j=0,\dots,4\).  Figure~\ref{fig:loss_matrix} visualizes the resulting \(4\times5\) “spectral‐loss” matrix: rows index the initial and four successive cut states, columns index the five fragment geometries (shown below).  Darker entries (lower values) indicate stronger intrinsic shape similarity.  Notice that for every state \(s_i\), the piece with low spectral distance in previous state still remain in small value despite the fact that the piece might fall down due to gravity and there is a new piece with low spectral distance appears.  This demonstrates both the pose-invariance and discriminative power of our spectral descriptor, and justifies its use for automatic labeling and qualitative evaluation of MPPI rollouts.

\subsection{MPPI for Data Generation}\label{app:mppi}

To generate high–quality demonstrations for policy learning we adopt the \textit{Model Predictive Path Integral} (MPPI) sampling–based optimal control scheme ~\cite{williams2017mppi}.  
MPPI maintains a horizon–\(H\) open–loop action sequence 
\(\mathbf{U}_{0:H-1} \in \mathbb{R}^{H\times m}\) (with \(m=6\) for the knife pose) and repeatedly refines it by importance–sampling noisy roll–outs in the deformable–body simulator.  
For each iteration we draw \(K\) perturbations \(\{\boldsymbol{\epsilon}^{(k)}\}_{k=1}^{K}\sim\mathcal{N}(0,\Sigma)\) and integrate the dynamics forward, harvesting the terminal fragments \(X^{(k)}\) produced by each sampled cut.  
The return of a roll–out is the negative spectral discrepancy
\[
J^{(k)} \;=\; -\,R_{\mathrm{spec}}\!\bigl(X^{(k)},g\bigr)
              -\sum_{t=0}^{H-1}\!\!C\!\bigl(o_t^{(k)},a_t^{(k)}\bigr),
\]
where \(R_{\mathrm{spec}}\) is the pose‑invariant spectral reward defined previously, and \(C(\cdot)\) is a small quadratic control penalty that biases the knife towards smooth, energy–efficient motions.

Given the roll–out returns we form importance weights  
\(
w_k \;=\; \exp\!\bigl(-J^{(k)}/\lambda\bigr)
\)
with temperature \(\lambda = 1/\kappa\) and update the mean action sequence in closed form:
\[
\mathbf{U}_{0:H-1} \;\leftarrow\;
\mathbf{U}_{0:H-1} \;+\;
\frac{\sum_{k=1}^{K}w_k\,\boldsymbol{\epsilon}^{(k)}}{\sum_{k=1}^{K}w_k}.
\]
After \(n\) refinement iterations the first action \(\mathbf{U}_0\) is executed on the robot; the horizon is then shifted and replanning continues until the episode terminates.  

During each MPPI episode we log the full state–action–reward tuples  
\(\{o_t,a_t,R_{\mathrm{spec}}(X_t,g)\}_{t=0}^{T}\)
as well as intermediate mesh reconstructions and knife trajectories.  
We repeat the procedure for a diverse set of goal shapes and material parameters, producing \(\sim\!6{,}000\) labelled cuts that serve as expert demonstrations for subsequent behaviour–cloning.

To complement the automatically generated dataset we developed a light‑weight tele‑operation interface in which the operator steers the knife with a 6‑DoF mouse while discrete cutting commands are triggered via keyboard hot‑keys.  
All tele‑operated actions are replayed in simulation to obtain exact particle damage labels and the same spectral reward used by MPPI, ensuring perfect consistency between human and algorithmic demonstrations.  

Together, MPPI planning guided by spectral rewards and targeted tele‑operation yield a rich, high‑fidelity dataset that underpins robust policy learning across the wide range of scenarios posed by \textit{\myalgo}.


\subsection{Policy Learning in Details}

\subsubsection{Dynamics-Informed Perception Module}\label{app:perception}

To efficiently encode dense deformable object states for policy learning, we develop a dynamic topology prediction model, denoted as \(\mathcal{F}\), which predicts the future topological state conditioned on the current topological state and action. Formally, we model the dynamics as:
\begin{align}
    \mathcal{F}: (\text{topo}_t, a_t) \mapsto \text{topo}_{t+1},
\end{align}
where both \(\text{topo}_t\) and \(a_t\) are represented as point clouds:
\begin{align}
\text{topo}_t = \{ (\mathbf{x}_i, \mathbf{f}_i) \}_{i=1}^N, \quad a_t = \{ (\mathbf{x}_j, \mathbf{g}_j) \}_{j=1}^N,
\end{align}
with \(\mathbf{x}_i, \mathbf{x}_j \in \mathbb{R}^3\) denoting 3D coordinates and \(\mathbf{f}_i\), \(\mathbf{g}_j\) denoting associated point features: cluster labels for topology and binary segmentation masks for action, respectively.

Within \(\mathcal{F}\), we introduce a perception encoder \(\Phi\) that processes point clouds by jointly embedding the 3D coordinates and their corresponding features:
\begin{align}
    \mathbf{z}_t = \Phi(\mathbf{X}_t, \mathbf{F}_t),
\end{align}
where \(\mathbf{X}_t\) and \(\mathbf{F}_t\) are the stacked xyz coordinates and feature vectors, respectively. \(\Phi\) produces a point-wise embedding \(\mathbf{z}_t\) that captures spatial geometry, topological structure, and action intent. This embedding is later reused for downstream goal-conditioned policy learning.

\paragraph{Topological Representation.}
For \(\text{topo}_t\), the point features \(\mathbf{f}_i\) encode the cluster membership obtained from our particle-based damage tracking, represented as a one-hot vector over a maximum of 32 clusters. For \(a_t\), the point features \(\mathbf{g}_j\) are binary one-hot vectors indicating whether the point lies on the cutting surface (cut = 1, not cut = 0).

\paragraph{Learning Objective and Intuition.}
The dynamic topology model \(\mathcal{F}\) is trained to predict how the object's topology evolves under a given action. Specifically, given \((\text{topo}_t, a_t)\), it predicts the resulting topological configuration \(\text{topo}_{t+1}\). By doing so, \(\Phi\) learns rich, actionable particle-wise embeddings that encode both geometric and topological transformations, enabling robust downstream manipulation.

\paragraph{Preprocessing and Graph Construction.}
Assuming access only to noisy surface observations, we reconstruct the object's mesh using Marching Cubes (Warp~\cite{warp2022}), sample interior points to form volumetric point clouds, and downsample them using Farthest Point Sampling (FPS). A topology-aware graph is then constructed, where edges connect nearest neighbors within the same cluster, promoting learning of local topological consistency.

\paragraph{Model Architecture.}
The perception encoder \(\Phi\) consists of two stages: (i) a Graph Convolutional Network (GCN) to extract local features from the topology-aware graph, and (ii) a Graph Transformer that globally refines the features via self-attention. The output is a set of point-wise embeddings encoding spatial and topological context.

\paragraph{Training Loss.}
We jointly optimize \(\Phi\) with two complementary objectives: geometric consistency and topological structure prediction. The overall loss is defined as:
\begin{align}
    \mathcal{L} = \lambda_{\text{pos}} \mathcal{L}_{\text{pos}} + \lambda_{\text{topo}} \mathcal{L}_{\text{topo}},
\end{align}
where \(\lambda_{\text{pos}}\) and \(\lambda_{\text{topo}}\) are balancing weights.

For geometric consistency, we minimize the Chamfer Distance (CD), Earth Mover's Distance (EMD), and Hausdorff Distance(HD) between predicted and ground-truth point clouds:
\begin{align}
    \mathcal{L}_{\text{pos}} = \text{CD}(\hat{X}, X) + \text{EMD}(\hat{X}, X) + \text{HD}(\hat{X}, X)
\end{align}
where \(\hat{X}\) and \(X\) are predicted and ground-truth point sets.

For topological structure prediction, we treat clustering as a contrastive learning task, where only the relative membership between points matters. A Hungarian matching-based loss is applied:
\begin{align}
    \mathcal{L}_{\text{topo}} = \min_{\pi \in \mathcal{P}} \sum_{i,j} \ell(\pi(c_i), \hat{c}_j),
\end{align}
where \(\mathcal{P}\) is the set of bipartite matchings, \(c_i\) and \(\hat{c}_j\) are the ground-truth and predicted cluster assignments, and \(\ell(\cdot, \cdot)\) is a binary cross-entropy loss.

\paragraph{Training Data.}
The dynamic topology model \(\mathcal{F}\) is pretrained using state-action-state triplets \((\text{topo}_t, a_t, \text{topo}_{t+1})\) collected from the MPPI-based demonstration generation (Section~\ref{sec:mppi}). This rich dataset covers diverse materials, object geometries, and cutting scenarios, enabling the learned embeddings to generalize effectively across a wide range of deformable manipulation tasks.

\subsubsection{Particle-based Score-Entropy Discrete Diffusion Policy}\label{app:pddb}

With the pretrained perception encoder \(\Phi\) that captures geometric and topological information, we now train a goal-conditioned behavior cloning (BC) policy for robotic cutting. The policy operates over perception embeddings, taking as input the current object topology and the desired goal shape, and predicting the next cutting action.

Given a point cloud observation \(o_t = (\mathbf{X}_t, \mathbf{F}_t)\) representing the current object state, and a goal point cloud \(g = (\mathbf{X}_g, \mathbf{F}_g)\), we first obtain their embeddings:
\begin{align}
    \mathbf{z}_t = \Phi(\mathbf{X}_t, \mathbf{F}_t), \quad \mathbf{z}_g = \Phi(\mathbf{X}_g, \mathbf{F}_g),
\end{align}
where \(\mathbf{z}_t\) and \(\mathbf{z}_g\) are point-wise embeddings encoding geometry and topology.

The cutting action is represented as a binary segmentation \(\mathbf{a}_t \in \{0,1\}^N\), where each point is classified as cut (1) or not cut (0).

\paragraph{Policy Model: Conditional Score-Based Discrete Diffusion.}
Inspired by Score Entropy Discrete Diffusion (SEDD)~\cite{lou2024sedd}, we formulate action prediction as a conditional discrete denoising diffusion process over point-wise binary labels.

In the forward process, the clean action labels \(\mathbf{a}_t^*\) are progressively corrupted into noisy labels \(\tilde{\mathbf{a}}_t\) through a multinomial noise distribution:
\begin{align}
    q_t(\tilde{\mathbf{a}}_t \mid \mathbf{a}_t^*) = \text{Multinomial}\left(\tilde{\mathbf{a}}_t \mid \mathbf{p}_t(\mathbf{a}_t^*)\right),
\end{align}
where \(\mathbf{p}_t(\mathbf{a}_t^*)\) denotes the noise schedule at timestep \(t\).

The policy network \(s_\theta\) is trained to predict the score function:
\begin{align}
    s_\theta(\tilde{\mathbf{a}}_t, t, \mathbf{z}_t, \mathbf{z}_g) \approx \nabla_{\tilde{\mathbf{a}}_t} \log q_t(\mathbf{a}_t^* \mid \tilde{\mathbf{a}}_t),
\end{align}
where \(s_\theta\) outputs the gradient of the log-posterior of the clean action given the noised action, conditioned on the current and goal embeddings.

This formulation enables the policy to iteratively denoise \(\tilde{\mathbf{a}}_t\) toward recovering the target cutting action.

\paragraph{Training Objective.}
The BC policy is trained by minimizing the Denoising Score Entropy (DSE) loss across all diffusion steps:
\begin{align}
    \mathcal{L}_{\text{BC}}(\theta) = \mathbb{E}_{(o_t, g, \mathbf{a}_t^*) \sim D} \left[ \sum_{t=1}^T \mathcal{L}_{\text{DSE}}(\tilde{\mathbf{a}}_t, \mathbf{a}_t^*; \theta) \right],
\end{align}
where the per-step DSE loss is defined as:
\begin{align}
    \mathcal{L}_{\text{DSE}}(\tilde{\mathbf{a}}_t, \mathbf{a}_t^*; \theta) = \mathbb{E}\left[\left\|s_\theta(\tilde{\mathbf{a}}_t, t, \mathbf{z}_t, \mathbf{z}_g) - \nabla_{\tilde{\mathbf{a}}_t} \log q_t(\mathbf{a}_t^* \mid \tilde{\mathbf{a}}_t)\right\|_2^2\right].
\end{align}

\paragraph{Action Reconstruction.}
After completing the denoising sampling process, we obtain a binary segmentation over the points indicating which regions should be cut. From the points classified as cut, we fit a cutting plane, thereby reconstructing the knife's pose and producing the final action \(a_t\).

\paragraph{1. Inputs, Goals, and Perception Embeddings}
At each decision step \(t\), the robot observes:
\[
o_t = (\mathbf{X}_t, \mathbf{F}_t),
\]
where:
\begin{itemize}
  \item \(\mathbf{X}_t = [\mathbf{x}_{t,1}, \ldots, \mathbf{x}_{t,N}]^\top \in \mathbb{R}^{N \times 3}\) are the 3D coordinates of \(N\) particles representing the object.
  \item \(\mathbf{F}_t = [\mathbf{f}_{t,1}, \ldots, \mathbf{f}_{t,N}]^\top \in \mathbb{R}^{N \times f}\) are associated per‐point features, such as normals, damage indicators, or material properties.
\end{itemize}
A desired goal shape is provided as another point‐cloud:
\[
g = (\mathbf{X}_g, \mathbf{F}_g) \in \mathbb{R}^{N \times 3} \times \mathbb{R}^{N \times f}.
\]
Both \(o_t\) and \(g\) are passed through a pretrained perception encoder
\[
\Phi: \bigl(\mathbb{R}^{N \times 3}, \mathbb{R}^{N \times f}\bigr) \;\longrightarrow\; \mathbb{R}^{N \times d},
\]
which uses graph‐based convolutions followed by a small MLP to produce point‐wise latent embeddings:
\[
\mathbf{z}_t = \Phi(\mathbf{X}_t, \mathbf{F}_t), 
\quad
\mathbf{z}_g = \Phi(\mathbf{X}_g, \mathbf{F}_g),
\]
where \(d\) is the embedding dimension.  Intuitively, \(\mathbf{z}_t\) captures the current object’s geometric and topological state, while \(\mathbf{z}_g\) encodes the desired target shape.

\paragraph{2. Action Representation and Forward Noising}
We represent the cutting action at time \(t\) as a binary mask
\[
\mathbf{a}_t^* = [a_{t,1}^*,\ldots,a_{t,N}^*]^\top \in \{0,1\}^N,
\]
where \(a_{t,i}^*=1\) indicates that particle \(i\) should be cut.  To train our policy using denoising diffusion, we first define a forward noising process that corrupts \(\mathbf{a}_t^*\) into progressively noisier versions \(\tilde{\mathbf{a}}_1,\dots,\tilde{\mathbf{a}}_T\).  Starting from
\(\tilde{\mathbf{a}}_0 = \mathbf{a}_t^*\), each step applies a small random flip:
\[
q_t\bigl(\tilde{\mathbf{a}}_t \mid \tilde{\mathbf{a}}_{t-1}\bigr)
=
\prod_{i=1}^N
\Bigl[
(1 - \beta_t)\,\delta\bigl(\tilde{a}_{t,i} = \tilde{a}_{t-1,i}\bigr)
\;+\;
\beta_t\,\delta\bigl(\tilde{a}_{t,i} \neq \tilde{a}_{t-1,i}\bigr)
\Bigr],
\]
where \(\beta_t\in[0,1]\) is a noise‐rate schedule (e.g.\ \(\beta_t = t/T\)).  After \(T\) steps, \(\tilde{\mathbf{a}}_T\) is nearly uniform random.

\paragraph{3. Score Network and Reverse Diffusion}
We train a neural network \(s_\theta\bigl(\tilde{\mathbf{a}}_t,\,t,\,\mathbf{z}_t,\,\mathbf{z}_g\bigr)\) to predict the gradient of the log‐probability of the true mask given the noisy mask:
\[
s_\theta(\cdot)
\;\approx\;
\nabla_{\tilde{\mathbf{a}}_t}\;\log\;
q_t\bigl(\mathbf{a}_t^* \mid \tilde{\mathbf{a}}_t\bigr).
\]
Concretely, \(s_\theta\) receives as input:
\begin{itemize}
  \item The noisy mask \(\tilde{\mathbf{a}}_t\) (embedded as scalars).
  \item The timestep \(t\), encoded with sinusoidal features.
  \item The concatenated perception embeddings \([\mathbf{z}_t; \mathbf{z}_g]\in\mathbb{R}^{N\times 2d}\).
\end{itemize}
During inference, we reverse the noising chain by sampling
\[
p_\theta\bigl(\mathbf{a}_{t-1} \mid \tilde{\mathbf{a}}_t, t, \mathbf{z}_t, \mathbf{z}_g\bigr)
=
\mathrm{Categorical}\Bigl(
\mathrm{softmax}\bigl(\log(1-\beta_t,\,\beta_t) + s_\theta(\tilde{\mathbf{a}}_t,t,\mathbf{z}_t,\mathbf{z}_g)\bigr)
\Bigr).
\]
This step “denoises” \(\tilde{\mathbf{a}}_t\) one level at a time back toward a clean mask.

\paragraph{4. Training Objective: Denoising Score‐Entropy Loss}
We optimize \(\theta\) by minimizing the expected squared error between the network’s predicted score and the true score over all timesteps:
\[
\mathcal{L}_{\mathrm{BC}}(\theta)
=
\mathbb{E}_{(o_t,g,\mathbf{a}_t^*)\sim D}
\Biggl[
\sum_{t=1}^T
\underbrace{
\mathbb{E}_{\tilde{\mathbf{a}}_t\sim q_t(\cdot\mid\mathbf{a}_t^*)}
\bigl\|\,
s_\theta(\tilde{\mathbf{a}}_t,t,\mathbf{z}_t,\mathbf{z}_g)
\;-\;
\nabla_{\tilde{\mathbf{a}}_t}\,\log\,q_t(\mathbf{a}_t^*\mid\tilde{\mathbf{a}}_t)
\bigr\|_2^2
}_{\mathcal{L}_{\mathrm{DSE}}(\tilde{\mathbf{a}}_t,\mathbf{a}_t^*;t)}
\Biggr].
\]
Because \(q_t\) uses simple bit‐flip noise, the true score gradient has a closed form:
\[
\nabla_{\tilde{a}_{t,i}}\log q_t(a_{i}^* \mid \tilde{a}_{t,i})
=\frac{\delta(\tilde{a}_{t,i}=a_{i}^*) - (1-\beta_t)}{\beta_t(1-\beta_t)}.
\]

\paragraph{5. Inference and Action Reconstruction}
At test time:
\begin{enumerate}
  \item Initialize \(\tilde{\mathbf{a}}_T\) by sampling each bit from \(\mathrm{Bernoulli}(0.5)\).
  \item For \(t=T, T-1, \ldots, 1\), sample \(\mathbf{a}_{t-1}\) from \(p_\theta(\mathbf{a}_{t-1}\mid \tilde{\mathbf{a}}_t,t,\mathbf{z}_t,\mathbf{z}_g)\).
  \item The final mask \(\tilde{\mathbf{a}}_0\) indicates the cut points.
  \item Fit a planar cut by solving
    \[
      \min_{\mathbf{n},d}\;
      \sum_{i:\,\tilde{a}_{0,i}=1}
      \bigl(\mathbf{n}^\top\mathbf{x}_{t,i} + d\bigr)^2,
      \quad
      \|\mathbf{n}\|_2 = 1,
    \]
    where \(\mathbf{n}\) is the plane normal and \(d\) its offset.
\end{enumerate}

\paragraph{Summary.}
Our BC policy systematically integrates dynamics-informed perception embeddings and conditional score-based discrete diffusion modeling. By treating cutting action prediction as a goal-conditioned denoising process over discrete labels, our method achieves robust and generalizable cutting behavior across diverse object shapes and cutting goals.

PDDP naturally handles the combinatorial nature of cutting (multimodal mask distributions) and provides smooth trade‐offs between sample quality and runtime via \(T\).  The closed‐form score supervision ensures stable training, and the iterative denoising generates crisp, coherent cutting actions that generalize across object shapes and materials.

Overall, our method systematically combines efficient topology tracking, pose-invariant shape evaluation, dynamics-informed perception, and supervised policy learning to enable effective robotic cutting in general scenarios.


\end{document}